\documentclass[11pt]{article}

\usepackage{times}
\usepackage{float}

\usepackage{fullpage}
\usepackage[compact]{titlesec}  
\titlespacing{\section}{0pt}{0pt}{0pt}
\titleformat{\section}
  {\normalfont\fontsize{13}{15}\bfseries}{\thesection}{1em}{}
\titleformat{\section}
  {\normalfont\fontsize{13}{15}\bfseries}{\thesection}{1em}{}
\usepackage[hidelinks]{hyperref}

\usepackage[utf8]{inputenc}
\usepackage{booktabs}
\usepackage{subcaption}
\usepackage{caption}
\usepackage{enumitem}
\usepackage{authblk}
\usepackage{caption}
\usepackage{multirow}
\usepackage{wrapfig}
\usepackage[round]{natbib}
\usepackage{textcomp}
\usepackage{adjustbox}
\usepackage{tabularx}
\usepackage{todonotes}
\usepackage{footnote}
\usepackage{graphicx}
\usepackage{fancyvrb} % for "\Verb" macro
\usepackage{hyperref}
\hypersetup{linkcolor=blue}
\usepackage{listings}
\usepackage{placeins}
\usepackage{amsmath}
\usepackage{amssymb}
\usepackage{pifont}
\usepackage{comment}

\newcommand{\ours}{Nemotron-4}

\title{Nemotron-4 15B Technical Report}
\author{Jupinder Parmar\footnote{
Equal contribution, corresponding authors: {\tt 
\{jupinderp,sprabhumoye,jjennings,mpatwary\}@nvidia.com}. 
} \ \ Shrimai Prabhumoye$^{*}$ \ \ Joseph Jennings$^{*}$ \ \ Mostofa Patwary$^{*}$ \\ Sandeep Subramanian\footnote{Work done while at NVIDIA.} \ \ Dan Su \ \ Chen Zhu \ \ Deepak Narayanan \ \  Aastha Jhunjhunwala \ \  Ayush Dattagupta \ \ Vibhu Jawa \ \ Jiwei Liu \ \ Ameya Mahabaleshwarkar \ \ Osvald Nitski \ \  Annika Brundyn \ \ James Maki \ \ Miguel Martinez \ \ Jiaxuan You \ \ John Kamalu \ \ Patrick LeGresley \ \ Denys Fridman \ \ Jared Casper \ \  Ashwath Aithal \ \ Oleksii Kuchaiev \ \ Mohammad Shoeybi \ \ Jonathan Cohen \ \ Bryan Catanzaro \par \textbf{NVIDIA}}

\date{}

\begin{document}

\parindent 0.0in
\parskip 0.15in

\maketitle

\begin{abstract}
We introduce \ours{} 15B, a 15-billion-parameter large multilingual language model trained on 8 trillion text tokens. \ours{} 15B demonstrates strong performance when assessed on English, multilingual, and coding tasks: it outperforms all existing similarly-sized open models on 4 out of 7 downstream evaluation areas and achieves competitive performance to the leading open models in the remaining ones. Specifically, \ours{} 15B exhibits the best multilingual capabilities of all similarly-sized models, even outperforming models over four times larger and those explicitly specialized for multilingual tasks.

%English, multilingual, and coding performance as it either outperforms, or is competitive to, all similarly-sized, open model in downstream benchmark tasks. Specifically, \ours{} 15B exhibits the best multilingual capabilities of all similarly-sized models, even outperforming models specialized for multilingual tasks.

% Reasoning: LM-EVAL: yes
% MMLU: no
% BBH" yes
% Coding: no
% Multilingual Classification : yes
% Multilingual Generation: yes
% Math: no

%We release \ours{} under the Apache 2.0 license.
%We introduce Nemotron-4 15B, a 15 billion parameter large language model trained on large amounts of high quality data and  demonstrates exceptional performance across benchmark tasks.
\end{abstract}

\section{Introduction}

Recently published efforts \citep{hoffmann2022training,touvron2023llama,touvron2023llama2,yang2023baichuan, jiang2023mistral} in language model pre-training have been inspired by Chinchilla scaling laws \citep{hoffmann2022training}, which argue for scaling data along with model size given a fixed compute budget, compared to past work that only scaled the size of the model~\citep{kaplan2020scaling,brown2020language, smith2022,rae2022scaling,workshop2023bloom}.
For example, \citep{hoffmann2022training} shows that given two roughly IsoFLOP GPT models with a similar data distribution, a 65-billion-parameter model on 1.4 trillion tokens and a 280-billion-parameter model on 300 billion tokens, the 65B model has better accuracy on downstream tasks.

This trade-off of allocating compute towards training on more data as opposed to increasing model size is particularly appealing from an inference perspective, reducing latency and the amount of compute needed to serve models. As a consequence, a major focus of language modeling training efforts has shifted to collecting high-quality multi-trillion token datasets from public sources such as Common Crawl. We continue this trend by introducing \ours{} 15B which was trained on 8 trillion tokens of English, multilingual, and coding text and was developed to be the best general-purpose large language model (LLM) that can fit on a single NVIDIA A100 or H100 GPU.
%node of NVIDIA A100s.  

%\ours{} 15B has been released under CC license to facilitate research and further customization of NeMoTron models.

\begin{figure}
\centering
\begin{subfigure}{.5\textwidth}
  \centering
  \includegraphics[width=\linewidth]{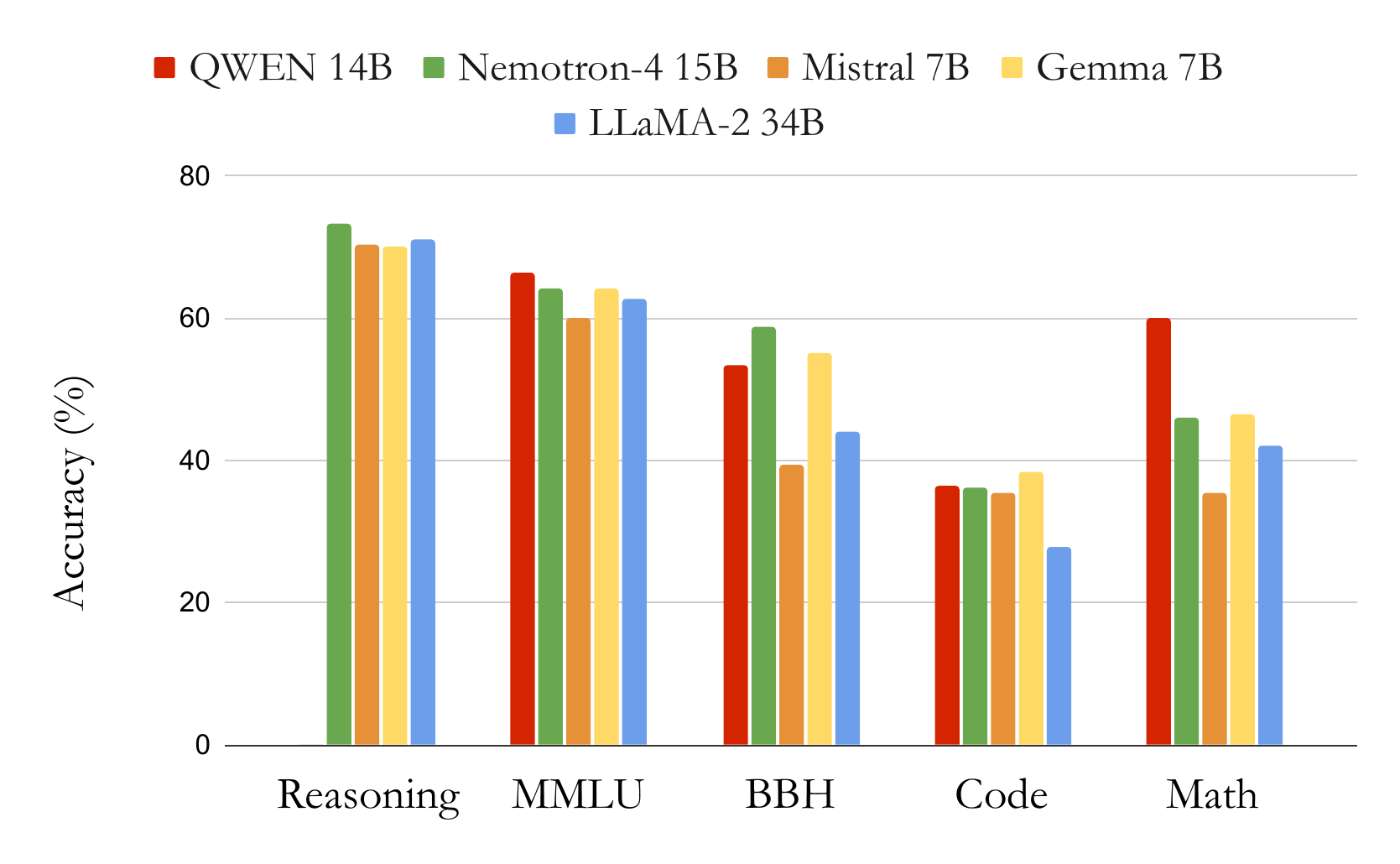}
  \label{fig:sub1}
\end{subfigure}%
\begin{subfigure}{.5\textwidth}
  \centering
  \includegraphics[width=\linewidth]{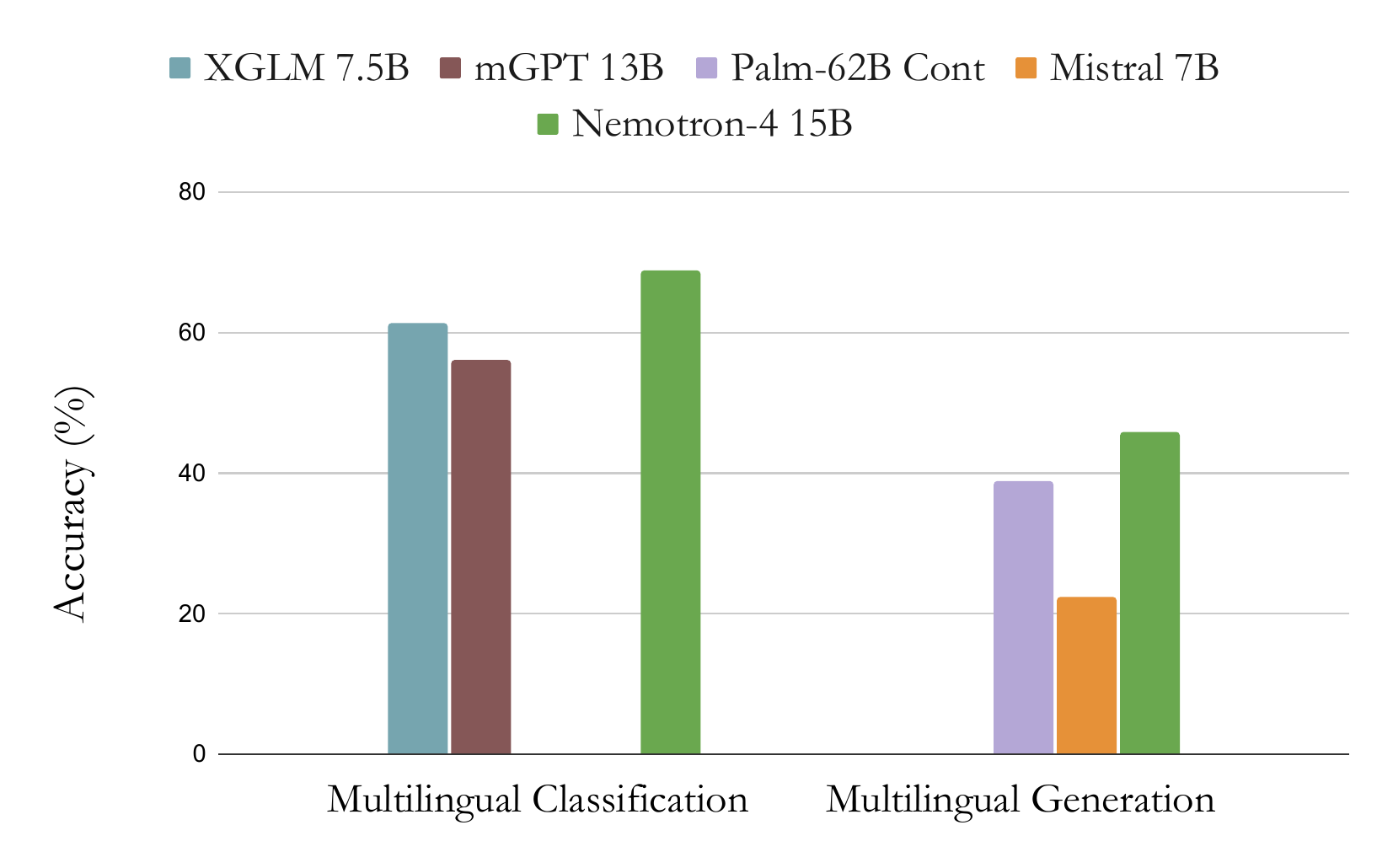}
  \label{fig:sub2}
\end{subfigure}
\caption{Comparison of \ours{} 15B across seven evaluation areas against similarly sized models. The composition of tasks that form each evaluation area can be found, along with more detailed evaluation results, in Section \ref{results}}
\label{fig:benchmark_comp}
\end{figure}

As demonstrated in Figure \ref{fig:benchmark_comp}, \ours{} 15B exhibits high downstream accuracies across a wide range of English, code, and multilingual evaluation areas. In comparison to leading similarly-sized, open models we show that \ours{} 15B is significantly better than LLaMA-2 34B~\citep{touvron2023llama2}, which has over twice the number of parameters, and is better than Mistral 7B~\citep{jiang2023mistral} on all English evaluation areas. Additionally, \ours{} 15B achieves competitive accuracies to QWEN 14B~\citep{bai2023qwen} and Gemma 7B~\citep{gemma24}. In a comparison across a wide range of programming languages, we find that \ours{} 15B achieves better average accuracy, and in particular on low-resource programming languages, than Starcoder~\citep{li2023starcoder}, a code-specific model, and Mistral 7B. As \ours{} 15B was trained on significant amount of multilingual data, it is currently the state-of-the-art general purpose model in its size class on all multilingual benchmarks. We find that \ours{} is better than PALM 62B-Cont~\citep{palm2}, and also outperforms multilingual-specific models such as XGLM~\citep{lin2022fewshot} and mGPT~\citep{shliazhko2022mgpt}.

%QWEN 14B~citep{qwen}
% TODO: chart  / table here to compare and show
%is competitive with all existing similarly-sized, open models on a variety of downstream English and coding evaluation areas.

%\begin{figure}
%    \centering
%    \includegraphics[width=0.85\linewidth]{comparison.png}
%    \caption{Data composition of the 8 trillion tokens used for pre-training.}
%    \label{fig:data_comp}
%\end{figure}

\begin{table}[h]
    \centering
    \begin{tabular}{ccccccc}
    \toprule
    Number of          & Hidden    & Number of       & Number of & Sequence & Vocabulary \\
    transformer layers & dimension & attention heads &    KV heads                & length   & size \\
    \toprule
    32 & 6144 & 48 & 8 & 4096 & 256,000 \\
    \bottomrule
    \end{tabular}
    \caption{Key hyper-parameters affecting size of \ours{} 15B.}
    \label{tab:model_arch}
\end{table}

\section{Architecture Details}

\ours{} uses a standard decoder-only Transformer architecture~\citep{DBLP:journals/corr/VaswaniSPUJGKP17}, with causal attention masks. Exact hyper-parameters affecting size are shown in Table~\ref{tab:model_arch}. 
\ours{} has 3.2 billion embedding parameters and 12.5 billion non-embedding parameters.
We use Rotary Position Embeddings (RoPE)~\citep{su2021roformer}, SentencePiece tokenizer~\citep{kudo2018sentencepiece}, squared ReLU activations in the MLP layers, no bias terms, dropout rate of zero, and untied input-output embeddings.
% Using untied embeddings also avoids an extra gradient all-reduce to keep embedding weights in sync across pipeline stages for models that use pipeline parallelism.
We use grouped query attention (GQA)~\citep{ainslie2023gqa} for faster inference and lower memory footprint.

\paragraph{Data.} 

We train \ours{} 15B on a pre-training dataset consisting of 8 trillion tokens. At a high-level, the data blend is split into three different types of data: English natural language data (70\%), multilingual natural language data (15\%), and source-code data (15\%). 
%Figure~\ref{fig:data_comp} highlights the distribution of the entire 8T token pre-training dataset.

\begin{figure}
    \centering
    \includegraphics[width=0.85\linewidth]{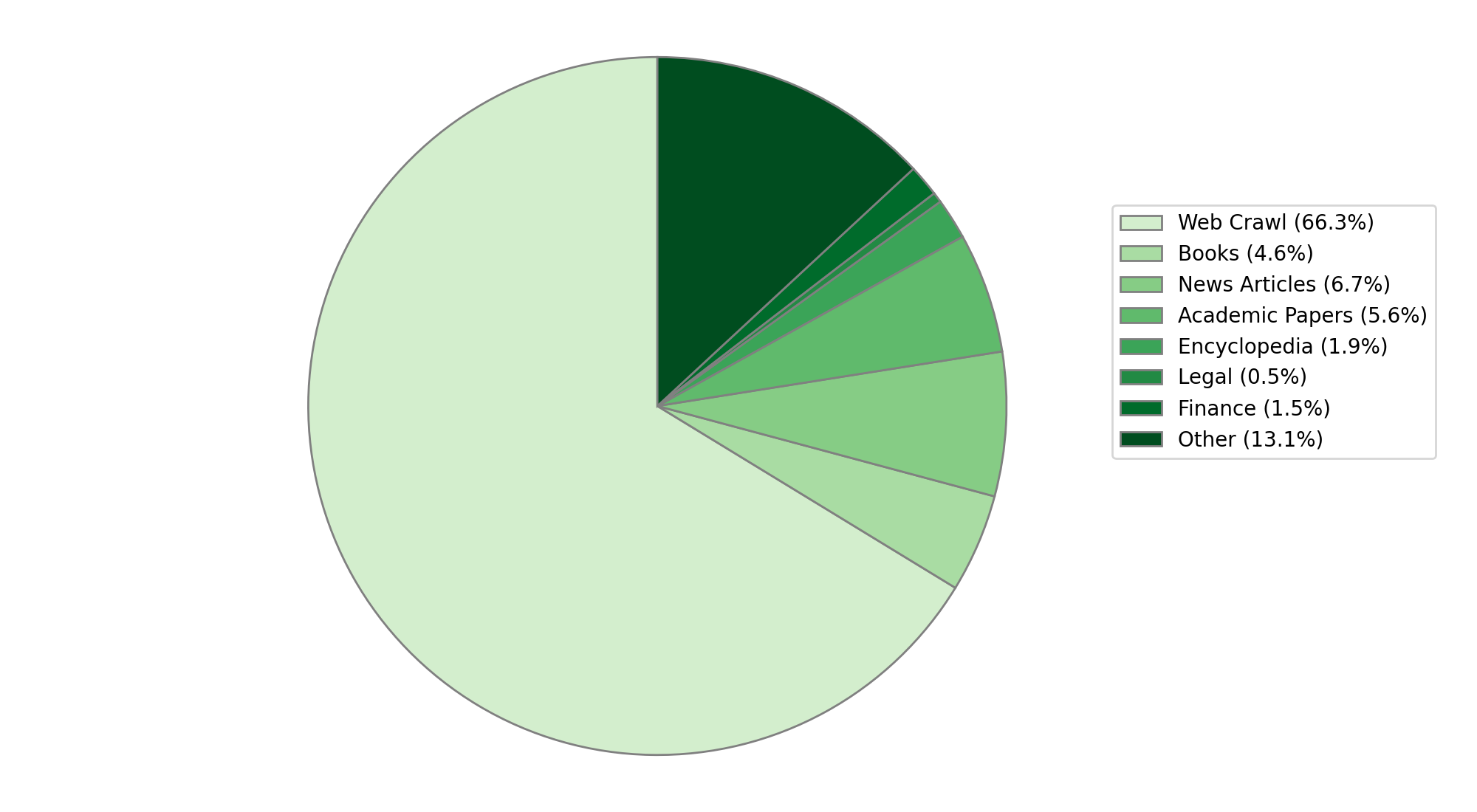}
    \caption{Data composition of the English tokens used for pre-training}
    \label{fig:data_comp}
\end{figure}

\begin{figure}[t]
    \centering
    \includegraphics[width=\linewidth]{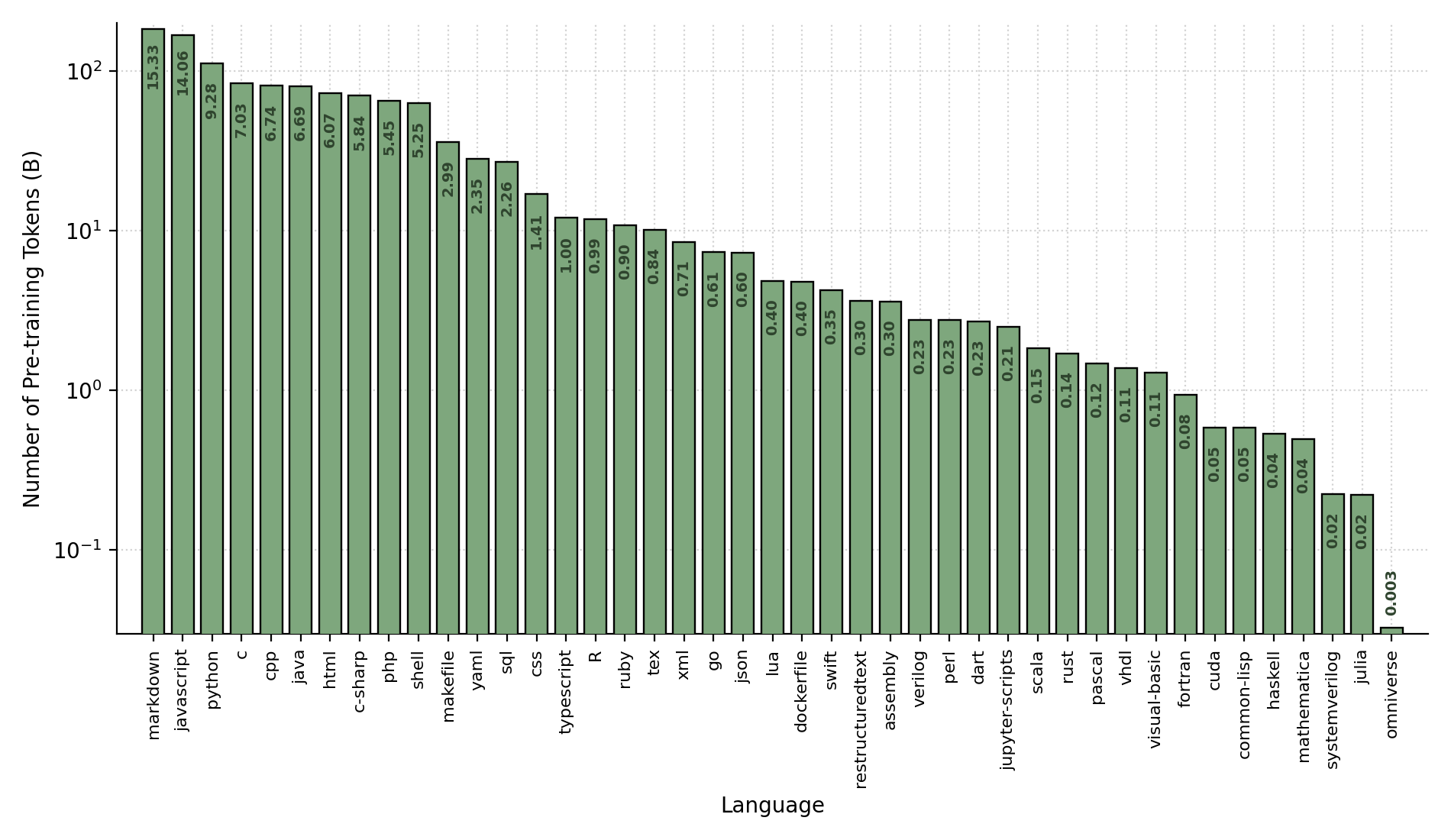}
    \caption{Data distribution of the 43 programming languages used for pre-training. The number within each bar indicates the percent of the overall code distribution that an individual language comprises.}
    \label{fig:code_distr}
\end{figure}

\begin{figure}[t]
    \centering
    \includegraphics[width=\linewidth]{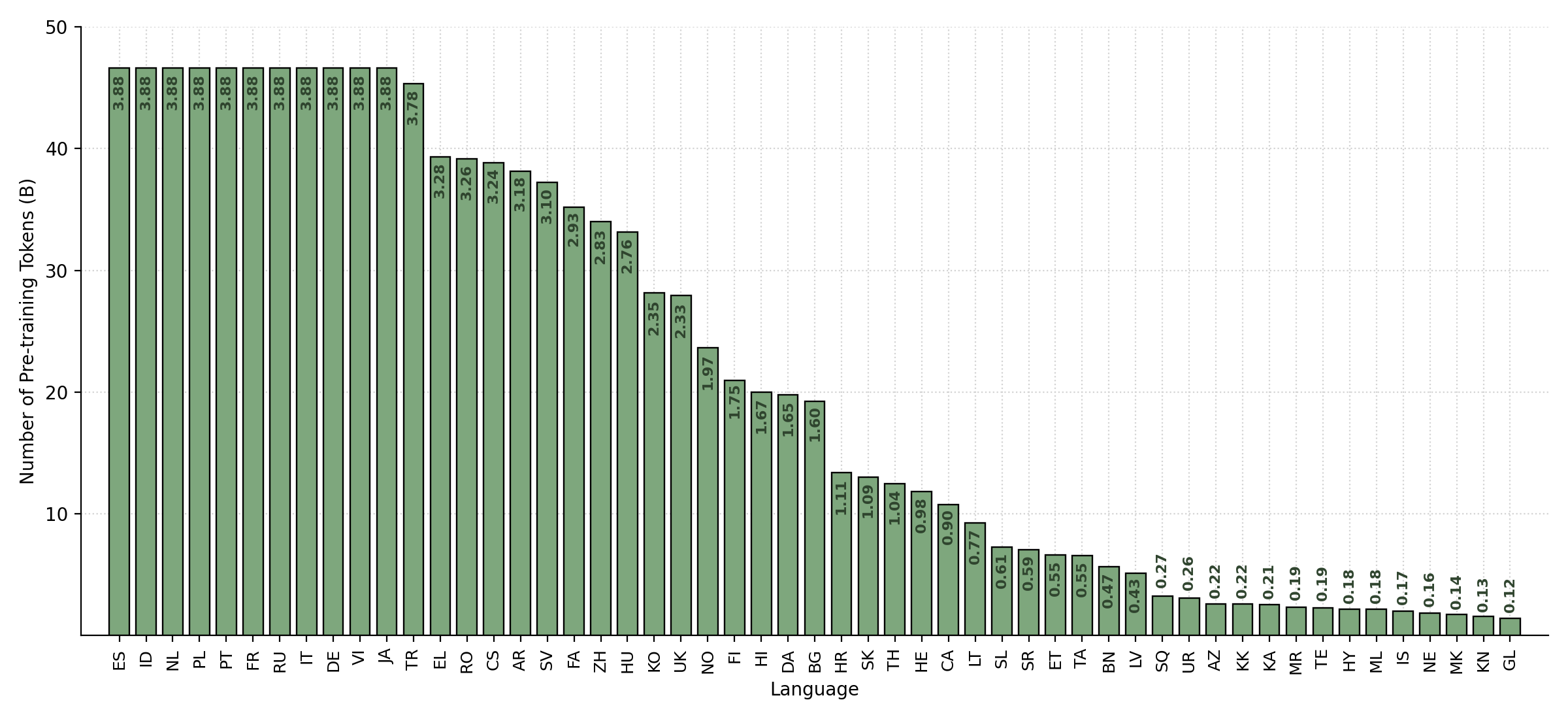}
    \caption{Data distribution of the 53 natural languages, aside from English,we used for pre-training. The number within each bar indicates the percent of the overall multilingual distribution that an individual language comprises.}
    \label{fig:multi_distr}
\end{figure}

The English corpus consists of curated documents from a variety of sources and domains including web documents, news articles, scientific papers, books, etc and the distribution used in our pre-training set is highlighted in Figure~\ref{fig:data_comp}. The code and multilingual data consists of a diverse set of natural and programming languages. We find that appropriately sampling tokens from these languages is key to strong accuracies in these domains. We share the distributions used for both code and multilingual tokens in our pre-training dataset in Figure~\ref{fig:code_distr} and Figure~\ref{fig:multi_distr} respectively.  

In constructing the pre-training corpus, we remove any possible duplicates via document-level exact and near-deduplication~\citep{jennings2023ndc}. We additionally applied document-level quality filtering across our corpus using a language-model based filtering approach similar to \citep{wenzek2019ccnet} in addition to a  series of heuristic filters as described in \citep{rae2022scaling} and \citep{raffel2020exploring}.

We train a BPE tokenizer in SentencePiece~\citep{kudo2018sentencepiece} on data that is randomly sampled from the final 8T token dataset. To have better coverage of low-resource languages in the tokenizer, we upsample non-English data relative to the final training dataset distribution. Our tokenizer preserves whitespaces (including leading and trailing ones), splits numbers into their individual digits~\citep{chowdhery2022palm}, and relies on byte-level backoff to handle unknown character sequences. The final vocabulary size is 256,000 tokens.

%\begin{figure}
%    \centering
%5    \includegraphics[width=0.7\linewidth]{multilingual.png}
%    \caption{Data composition of 8 trillion tokens used for pre-training}
%    \label{fig:data_comp}
%\end{figure}

\paragraph{Pre-training.} 
\ours{} was trained using 384 DGX H100 nodes; each node contains 8 H100 80GB SXM5 GPUs based on the NVIDIA Hopper architecture~\citep{nvidia2022h100}. 
Each H100 GPU has a peak throughput of 989 teraFLOP/s when doing 16-bit floating point (\texttt{bfloat16}) arithmetic without sparsity. Within each node, GPUs are connected by NVLink and NVSwitch~\citep{nvlink}; the GPU-to-GPU bandwidth is 900 GB/s (450 GB/s in each direction). Each node has 8 NVIDIA Mellanox 400 Gbps HDR InfiniBand Host Channel Adapters (HCAs) for inter-node communication. 
%The switch architecture is a three level fat tree topology (leaf, spine, and core) implemented using a total of 1200 switches. This switch architecture provides full port bandwidth between rail aligned HCAs.

%Each node also contains an additional 2 HCAs for access to the cluster wide Lustre filesystem. Peak storage bandwidth is 50 GB/s/node with a maximum of 2 TB/s across the entire cluster.

We used a combination of 8-way tensor parallelism~\citep{shoeybi2019megatron} and data parallelism to train the model; we also use a distributed optimizer to shard the optimizer state over the data-parallel replicas. The degree of data parallelism was varied from 96 to 384 as the batch size was ramped up. 
Table~\ref{tab:43bmodel_batch_time_mfu} summarizes the 3 stages of batch size ramp, and includes the per-iteration time and model FLOP/s utilization (MFU)~\citep{chowdhery2022palm,korthikanti2022sequence}. MFU quantifies how efficiently the GPUs are utilized in model training. Training was completed in approximately 13 calendar days.
%\shrimai{Overlapping computation and data parallel communication, and the high internode bandwidth of CoreWeave, results in almost perfectly linear data parallel scaling.}

%The last two columns are the product of the time per iteration and number of iterations at each batch size. 
%, concurrently with other workloads running on the system. 

\begin{table}[H]
    \centering
    \begin{tabular}{crccccc}
    \toprule
    Data-parallel size & GPUs & Iteration time (secs) & MFU (\%) & Batch size & Tokens (B) & Time (days) \\
    \toprule
    96 &  768 & 0.57 & 34.3 & 384 &  200 & 0.8 \\
    192 & 1,536 & 0.58 & 33.3 & 768 &  200 & 0.4 \\
    288 & 2,304 & 0.64 & 30.5 & 1,152 &  7,600 & 11.9   \\
    % 384 & 3072 & 0.61 & 24.2 & 1152 &  7600 & 11.4 \\
    \bottomrule
    \end{tabular}
    \caption{Batch size rampup schedule, along with time and efficiency metrics for the \ours{} 15B parameter model.}
    \label{tab:43bmodel_batch_time_mfu}
\end{table}

\paragraph{Continued Training.}
Similar to recent work~\citep{geminiteam2023gemini}, we find that switching the data distribution and learning rate decay schedule at the end of model training greatly improves model quality. Concretely, after having trained over the entirety of our 8T pre-training dataset, we use the same loss objective and perform continued training on small number of tokens in comparison to the pre-training tokens. 

In this additional phase of continued training, we utilize two distinct data distributions. 
The first distribution is where the majority of tokens during continued training are sampled from. It utilizes tokens that have already been introduced during pre-training but with a distribution that places larger sampling weight on higher quality sources. 
The second distribution introduces a small number of benchmark-style alignment examples to better allow the model to respond to such questions in downstream evaluations while also up-weighting data sources that come from areas of low model performance. 
In accompaniment with a learning rate schedule that prioritizes a steeper slope of decay than magnitude of learning rate, we find that such an ordering and style of data distributions allows for the model to gently transition from the pre-training dataset and better learn newly emphasized data areas.

\section{Results}\label{results}
We evaluate \ours{} 15B on a variety of downstream evaluation areas covering a diverse range of tasks and domains. In all evaluations, we adhere to the standardized task setup and share the exact settings used. The covered evaluation categories include: 

\begin{itemize}
    \item \textbf{Commonsense Reasoning (0-shot):}  SIQA \citep{sap2019socialiqa}, ARC easy and challenge \citep{clark2018think},  PIQA \citep{Bisk2020PIQARA}, Winogrande \citep{Sakaguchi2020WINOGRANDEAA}, and Hellaswag  \citep{Zellers2019HellaSwagCA}
    \item \textbf{Popular Aggregated Benchmarks:} MMLU (5-shot) \citep{hendrycks2020measuring} and BBH (3-shot) \citep{suzgun2022challenging}
    \item \textbf{Math:} GSM8K (8-shot with maj@1) \citep{cobbe2021}
    \item \textbf{Code:} Pass@1 scores on HumanEval (0-shot) \citep{chen2021evaluating}, MBPP (3-shot) \citep{austin2021program}, and MultiPL-E (0-shot) \citep{multiple2023}
    \item \textbf{Multilingual:} classification via XCOPA (0 and 4-shot) \citep{xcopa}, machine translation with FLORES-101 (8-shot)~\citep{goyal2021}, and generation tasks such as MGSM (8-shot)~\citep{shi2022language} and TyDiQA (1-shot)~\citep{tydiqa} 
\end{itemize}

%can add in trivia qa / truthful qa after this aggregation to see. -- confirm with shrimai and mostofa 

%We evaluate our models on a variety of benchmarks. These benchmarks include:
%\begin{itemize}
%    \item \textbf{English Benchmarks:} These tasks broadly test commonsense reasoning, reading comprehension, math and world knowledge. We evaluate on \textit{Language Model Evaluation Harness}~\citep{eval-harness}, \textit{BigBench}~\citep{srivastava2022beyond}, \textit{Massive Multitask Language Understanding} (MMLU)~\citep{hendrycks2020measuring} and Big-Bench hard ~\citep{srivastava2022beyond}.
%    \item \textbf{Multilingual Benchmarks:} classification tasks like XCOPA~\citep{xcopa}, machine translation tasks like FLORES-101~\citep{goyal2021}, and generation tasks like MGSM~\citep{shi2022language}, and TyDiQA~\citep{tydiqa}.
%    \item \textbf{Code Benchmarks:} HumanEval~\citep{chen2021evaluating}, MBPP~\citep{austin2021program} and Multiple-e~\citep{multiple2023}.
%\end{itemize}

% and toxicity generation evaluation following the RealToxicityPrompts setup. 
%These evaluations cover a variety of tasks: common sense reasoning, reading comprehension, question answering, natural language inference, and more. 

In our evaluations, we compare against a number of external decoder-only transformer language models and unless otherwise stated we use the numbers published in the reports of the corresponding models. For English and code tasks, we share detailed results for \ours{} 15B, LlaMA-2 13B and 34B \citep{touvron2023llama2}, Mistral 7B \citep{jiang2023mistral}, Baichuan-2 13B~\citep{yang2023baichuan}, QWEN 14B~\citep{bai2023qwen}, and Gemma 7B \citep{gemma24}. For multilingual benchmarks, we report results against PaLM 62B and 62B-cont \citep{chowdhery2022palm} as well as models specially trained for multilingual capabilities such as mGPT 13B \citep{shliazhko2022mgpt} and XGLM 7.5B \citep{lin2022fewshot}.

\subsection{Commonsense Reasoning}
We use the LM-Evaluation Harness~\citep{eval-harness} to evaluate \ours{} 15B across all aforementioned tasks. Table~\ref{tab:lm_eval_harness_external} showcases that \ours{} 15B achieves the strongest average performance on this diverse set of tasks.

\begin{table}[H]
\centering
\begin{adjustbox}{max width=\textwidth, scale=1.0}
\begin{tabular}{lrccccccc}
\toprule
& Size & SIQA  & ARC-c & ARC-e & PIQA & Winogrande & Hellaswag & AVG \\ 
\midrule
%GPT-NeoX  & 20B & 72.0 & 72.3 & - & - & 34.3 & 77.9 & 66.1 & 53.5 & 65.8 \\
%\midrule
%GPT-3 &  175B & 76.2 & 68.8 & 14.4 & 45.5 & 35.4 & 81.0 & 70.2 & 78.9 & 76.7 \\
%\midrule
%MT-NLG & 530B & 76.6 & - & - & 47.9 & 36.6 & 82.0 & 73.0 & 80.2 & 78.4 \\
%\midrule
%\multirow{4}{*}{PaLM} & 8B & 69.5 & 69.2 & 6.7 & 42.3 & 35.8 & 77.1 & 66.3 & 68.7 & 70.7\\
% & 62B & 75.4 & 75.2 & 11.5 & 47.5 & 37.2 & 80.5 & 77.0 & 79.7 & 79.1  \\
% & 62B-cont & 76.3 & - & - & - & - & 81.4 & 77.0 & 80.6 & \textbf{79.7} \\
%\midrule
\multirow{2}{*}{LLaMA-2} & 13B & 50.3 & 49.4 & 77.3 & 79.8 & 72.8 & 80.7 & 68.4 \\
& 34B & 50.9 & 54.5 & 79.4 & 81.9 & 76.7 & 83.3 & 71.1 \\
\midrule
\multirow{1}{*}{Baichuan-2} & 13B & - & - & - & 78.1 & - & 70.8 & - \\
\midrule
QWEN & 14B & 77.9 & 84.4 &90.3 & 79.9 & - & 80.2 & - \\
\midrule
Mistral  & 7B & \ \ 47.0$^{*}$ & 55.5 &80.0 & 83.0 & 75.3 & 81.3 & 70.4 \\
\midrule
Gemma & 7B & 51.8 & 53.2 & 81.5 & 81.2 & 72.3 & 81.2 & 70.2 \\
\midrule
\ours{} & 15B & 60.9 & 55.5 & 80.9 & 82.4 & 78.0 & 82.4 & \textbf{73.4} \\
\bottomrule
\end{tabular}
\end{adjustbox}
\caption{\label{tab:lm_eval_harness_external} Results on standard reasoning benchmarks in the zero-shot setting. We report the average across all tasks where possible for a fair comparison. The values marked with $*$ are read from ~\citet{gemma24} }
\end{table}

\begin{comment}
\begin{table}[H]
\centering
\begin{adjustbox}{max width=\textwidth, scale=1.0}
\begin{tabular}{lrccccccccc}
\toprule
& Parameters  & Lambada & ARC-Easy & WebQS & Race-H & ANLI-R2 & PiQA & Winogrande & Hellaswag & AVG \\ 
\midrule
%GPT-NeoX  & 20B & 72.0 & 72.3 & - & - & 34.3 & 77.9 & 66.1 & 53.5 & 65.8 \\
%\midrule
%GPT-3 &  175B & 76.2 & 68.8 & 14.4 & 45.5 & 35.4 & 81.0 & 70.2 & 78.9 & 76.7 \\
%\midrule
%MT-NLG & 530B & 76.6 & - & - & 47.9 & 36.6 & 82.0 & 73.0 & 80.2 & 78.4 \\
%\midrule
%\multirow{4}{*}{PaLM} & 8B & 69.5 & 69.2 & 6.7 & 42.3 & 35.8 & 77.1 & 66.3 & 68.7 & 70.7\\
% & 62B & 75.4 & 75.2 & 11.5 & 47.5 & 37.2 & 80.5 & 77.0 & 79.7 & 79.1  \\
% & 62B-cont & 76.3 & - & - & - & - & 81.4 & 77.0 & 80.6 & \textbf{79.7} \\
%\midrule
\multirow{2}{*}{LLaMA-2} & 7B & - & 75.2 & - & - & - & 78.8 & 69.2 & 77.2  & 75.1 \\
& 13B & - & 77.3 & - & - & - & 79.8 & 72.8 & 80.7 & 77.7 \\
& 34B & - & 79.4 & & & & 81.9 & 76.7 & 83.3 & 80.3 \\
\midrule
\multirow{2}{*}{Baichuan-2} & 7B & - & - & - & - & - & 76.2 & - & 67.0 & - \\
& 13B & - & - & - & - & - & 78.1 & - & 70.8 & - \\
\midrule
Mistral  & 7B & - & 80.0 & - & - & - & 83.0 & 75.3 & 81.3 & 79.9\\
\midrule
\ours{} & 15B & 75.7 & 80.9 & 22.3 & 47.6 & 44.8 & 82.4 & 78.0 & 82.4 & \textbf{80.9} \\
\bottomrule
\end{tabular}
\end{adjustbox}
\caption{\label{tab:lm_eval_harness_external}Zero-shot accuracy of models on many NLU tasks. We report the average over ARC-Easy,  PiQA, Winogrande, and Hellaswag where possible. }
\end{table}
\end{comment}

\subsection{Popular Aggregated Benchmarks}

The MMLU~\citep{hendrycks2020measuring} and Big Bench Hard (BBH)~\citep{suzgun2022challenging} benchmarks have been developed as a challenging assessment of language models' capabilities on a wide range of tasks and domains. As seen from  Table~\ref{tab:bbh_mmlu}, \ours{} 15B achieves the best score on BBH across existing models at its scale by nearly 7\%. 
Additionally, \ours{} is significantly better than LLaMA-2 70B model on BBH benchmark where LLaMA-2 70B attains a score of 51.2 and \ours{} is 58.7.
\ours{} 15B additionally attains a highly competitive MMLU score and its per-category performance on MMLU can be found in Table \ref{tab:mmlu}.

%\subsection{MMLU}
%\subsection{Bigbenchhard}

%\shrimai{put avg, maybe details in appendix}

%& 70B & 51.2 &  \\

\begin{table}[h]
\centering
\begin{tabular}{lrcc} 
\toprule
    & Size & BBH & MMLU     \\ 
\midrule
\multirow{3}{*}{LLaMA-2} & 13B  & 39.4 & 54.8 \\
& 34B & 44.1 & 62.6 \\
\midrule
\multirow{1}{*}{Baichuan-2} & 13B  & 48.8 & 59.2 \\
\midrule
QWEN & 14B & 53.4 & \textbf{66.3} \\
\midrule
Mistral & 7B & 39.5 & 60.1 \\
\midrule
Gemma & 7B & 55.1 & 64.3 \\
\midrule
\ours{} & 15B & \textbf{58.7} & 64.2 \\ 
\bottomrule
\end{tabular}
\caption{\label{tab:bbh_mmlu} \ours{} 15B attains highly competitive performance on popular aggregate benchmarks. The BBH result for Mistral is read from the figure in~\citep{jiang2023mistral}.}
\end{table}

\begin{comment}
\begin{table}[H]
\centering
\begin{tabular}{lccccc} 
\toprule
Sub-tasks & \ours{} &  &       \\ 
boolean\_expressions & \\
causal\_judgement  & \\
date\_understanding  & \\
disambiguation\_qa  & \\
dyck\_languages  & \\
formal\_fallacies  & \\
geometric\_shapes  & \\
hyperbaton  & \\
logical\_deduction\_five\_objects  & \\
logical\_deduction\_seven\_objects  & \\
logical\_deduction\_three\_objects  & \\
movie\_recommendation  & \\
multistep\_arithmetic\_two  & \\
navigate  & \\
object\_counting & \\
penguins\_in\_a\_table  & \\
reasoning\_about\_colored\_objects & \\
ruin\_names & \\
salient\_translation\_error\_detection & \\
snarks & \\
sports\_understanding & \\
temporal\_sequences & \\
tracking\_shuffled\_objects\_five\_objects & \\
tracking\_shuffled\_objects\_seven\_objects & \\
tracking\_shuffled\_objects\_three\_objects & \\
web\_of\_lies & \\
word\_sorting & \\
\bottomrule
\end{tabular}
\caption{\label{tab:bbh} Three-shot performance on the Big-Bench Hard.}
\end{table}
\end{comment}

\subsection{Math and Code}

Recently, large language models have been shown to be effective at both mathematical reasoning and a variety of coding tasks \citep{allal2023santacoder,chowdhery2022palm, touvron2023llama}. Table~\ref{tab:code_external} highlights the performance of \ours{} 15B on such tasks. Specifically, on mathematical reasoning we find that \ours{} 15B achieves strong performance as it attains a similar score to Gemma 7B, but lags behind models such as Baichuan-2 and QWEN. On code tasks, we see that \ours{} performs on par with QWEN 14B while remaining slightly behind Gemma 7B. Across both types of tasks, \ours{} 15B is able to outperform Mistral 7B and LlaMA-2 13B/34B.

%The p@1 numbers for both the tasks are reported in Table~\ref{tab:code_external}.
%As we see in Table~\ref{tab:code_external}, \ours{} performs better than LLaMA-2 34B and even LLaMA-2 70B on Humaneval task.
%It performs better than Mistral 7B and competitively with similar sized QWEN 14B model.
%For MBPP task, it performs better than LLaMA-2 13B and 34B models and it performs competitively with QWEN 14B and Gemma 7B.

%a variety of coding tasks including: code translation~\citep{chowdhery2022palm}, code completion \citep{allal2023santacoder,chowdhery2022palm, touvron2023llama}, and code infilling~\citep{bavarian2022efficient, nguyen2023meet}. 
%In this section, we show that \ours{} achieve exemplary coding performance on HumanEval~\citep{chen2021evaluating} and MBPP~\citep{austin2021program}.

%These two tasks are \textit{Text-to-Code} tasks where the objective is to write code which implements a specification given in natural language. 
%In both, the model is provided with a short natural-language description and a small set of input-output examples illustrating the desired output and it is tasked with generating a Python program that meets the requirement. 
% & 70B & 29.9 & \textbf{45.0} \\

\begin{table}[H]
\centering
\begin{adjustbox}{max width=\textwidth, scale=1.0}
\begin{tabular}{lrccc}
\toprule
& Size & GSM8K & HumanEval & MBPP \\
\midrule
  \multirow{2}{*}{LlaMA-2} & 13B & 28.7 & 18.3 &30.6 \\ 
 & 34B & 42.2 & 22.6 & 33.0 \\
 \midrule
 \multirow{1}{*}{Baichuan-2}  & 13B & 52.8 & 17.1  & 30.2 \\ 
 \midrule
 \multirow{1}{*}{QWEN}  & 14B & \textbf{60.1} & 32.2 & 40.8 \\
 \midrule
\multirow{1}{*}{Mistral}  & 7B & 35.4$^{*}$  &  30.5 &  \ \ 40.2$^{*}$   \\
\midrule
\multirow{1}{*}{Gemma}  & 7B & 46.4 &  \textbf{32.3} & \textbf{44.4} \\
\midrule
\multirow{1}{*}{\ours{}}  & 15B & 46.0 &   31.6 & 40.6  \\
\bottomrule
\end{tabular}
\end{adjustbox}
\caption{\label{tab:code_external} Comparative results on math and code benchmarks. As Mistral 7B reports MBPP performance on a different eval split and uses a different evaluation setting for GSM8K , we use the corresponding numbers reported in~\citep{gemma24}}
\end{table}

Nearly all similarly-sized open models determine their code abilities solely based on performance on Python related tasks -- disregarding an evaluation of their capabilities on other programming languages. In Table~\ref{tab:multiple-e}, we demonstrate results of \ours{} 15B on the Multiple-E~\citep{cassano2023multipl} benchmark across 11 diverse programming languages and compare it against Mistral 7B and Starcoder~\citep{li2023starcoder}, a 15B parameter model that has been specially trained for code. We find that \ours{} 15B attains strong coding performance across a wide assortment of programming languages and outperforms both Starcoder and Mistral 7B on average. We especially highlight the superior performance of \ours{} 15B on low-resource programming languages such as Scala, Julia, and R.

\begin{table}[t]
\centering
\begin{adjustbox}{max width=\textwidth, scale=1.0}
\begin{tabular}{lccccccccccccc}
\toprule
  & Size & JavaScript & Julia & Java & Lua & C++ & C-Sharp & PHP & Shell & TypeScript & R & Scala & AVG  \\
\midrule
Starcoder & 15B & 30.8 & 23.0 & 30.2 & 23.9 & 31.6 & 21.0 & 26.1	& 10.5 & 32.3 & 15.5 & 27.6 & 24.2 \\
Mistral & 7B & 34.2 & 22.0 & 26.0 & 25.3 & 29.1 & 22.8 & 27.9	& 8.9 & 28.5 & 11.8 & 22.2 & 23.6 \\
\ours{} & 15B & 28.6 & 24.8 & 24.8 & 24.2 & 35.4 & 21.1 & 27.3 & 8.9 & 32.9 & 18.6 & 27.3 & \textbf{24.5} \\ 
\bottomrule
\end{tabular}
\end{adjustbox}
\caption{\label{tab:multiple-e} \ours{} 15B attains high competency in coding performance across a broad range of programming languages. Results for Mistral are from our runs of Mistral in the same setting as \ours{}. }
\end{table}

\subsection{Multilingual}

We demonstrate the outstanding multilingual ability of \ours{} 15B using four widely-studied benchmarks in previous works that cover a diverse range of high to low resource natural languages. For classification we use accuracy as the metric; for generative tasks, we use exact match; and for machine translation, we evaluate using the \texttt{sacreBLEU} \citep{post2018} implementation of \texttt{BLEU}~\citep{papineni2002}, using \texttt{spm-flores-101} tokenization to obtain spBLEU scores. 

%In our generations, we use greedy decoding until either a \texttt{EOS} token is generated or the maximum number of desired tokens have been generated. 

\textbf{1. Classification:}  Cross-lingual Choice of Plausible Alternatives (XCOPA) \citep{xcopa} tests causal commonsense reasoning in 11 languages 
%abd was created by direct translation of the English Choice of Plausible Alternatives (COPA) \citep{copa} dataset. 
%In this evaluation, the objective is to correctly select which of two premises fits most plausibly with a possible hypothesis. 
%An example from XCOPA is shown in Figure \ref{fig:multilingual_examples}.

We compare \ours{} 15B to existing multilingual language models: XGLM \citep{lin2022fewshot} , mGPT \citep{shliazhko2022mgpt}, and BLOOM \citep{workshop2023bloom}. 
XGLM and mGPT are models specially trained to have improved multilingual ability by up-sampling the presence of non-English languages in the training data. 
In contrast, BLOOM, like \ours{}, is a general purpose language model that was trained on a combination of English, multilingual, and code data. 
In Table \ref{tab:xcopa}, we clearly see that \ours{} achieves the best performance amongst all models -- realizing almost a 12\% improvement in the four-shot setting. 

%Overall, \ours{} clearly beats BLOOM, a general purpose language model with over 11$\times$ the number of parameters, and consistently outperforms across the zero and few shot settings against two models specialized for multilingual tasks. 
%Specifically, we see a gain of 6\% in zero-shot setting and 23\% in four-shot setting over the similar-sized mGPT model.

\begin{table}[H]
\centering
\begin{adjustbox}{max width=\textwidth, scale=1.0}
\begin{tabular}{lcrcccccccccccccc}
\toprule
Mode & Model & Size & ET & HT & ID & IT  & QU & SW & TA  & TH & TR & VI & ZH & AVG \\
\midrule
 \multirow{4}{*}{Zero-Shot} & BLOOM &   176B & - & - &  $57.5^*$ &	- &	 -  &	$59.5^*$ &	 $54.7^*$ &	- &	- &  $58.2^*$ &  $57.7^*$ & -     \\
 & XGLM & 7.5B & 57.6	& 57.0	& 59.0 &	49.2 &	52.4&	55.0 &	55.6	& 57.8& 	55.0 &	59.0 &	53.6 &	55.6 \\
 & mGPT &   13B & 49.8 &	50.4	& 63.4	& 61.6	& 50.4	& 57.6 &	57.0 & 54.0 &	58.2	& 60.4	& 54.6 & 56.1  \\
  & \ours{} & 15B & 62.8 & 47.4 & 66.6 & 67.0 & 53.8 & 50.4 & 62.0 & 59.6 & 57.4 & 65.2 & 62.2 & \textbf{59.5}   \\
  \midrule
\multirow{3}{*}{4-Shot} & XGLM &   7.5B & 64.7	&60.4	&67.3	& 64.0&	50.0&	61.8&	56.7&	61.5&	60.1 &	68.5 & 59.9	& 61.4   \\
 & mGPT & 13B & 48.6&	48.6	&62.6&	60.8&	50.6&	56.6&	55.4&	54.8&	57.4	&61.8&	58.4&56.0            \\
 & \ours{} & 15B & 72.9 & 52.8 & 79.6 & 79.2 & 50.2 & 52.2 & 72.8 & 66.6 & 77.2 & 78.6 & 76.0 & \textbf{68.9}  \\
\bottomrule
\end{tabular}
\end{adjustbox}
\caption{\label{tab:xcopa} Comparison of \ours{} 15B against existing large language models on XCOPA under the zero- and four-shot setting. Our reported results for XGLM are from the runs of the model in \citep{shliazhko2022mgpt} given that we use the same prompt template used by mGPT. The values marked with $*$ are read from figures in \citep{workshop2023bloom}. %The average of the zero-shot results is only over ID, SW, TA, VI, and ZH whereas the four-shot average is across all languages. \shrimai{check 0-shot avg} 
}
\end{table}

\textbf{2. Generation:} We consider two generative tasks: TyDiQA-GoldP~\citep{tydiqa} and Multilingual Grade School Math (MGSM)~\citep{shi2022language}. TyDiQA-GoldP is a question answering task while MGSM evaluates the arithmetic reasoning ability of language models in 10 languages.

%and was created through direct translation of 250 examples from the GSM8K benchmark. 
%MGSM is particularly useful as it has been shown to highlight emergent abilities within language models. 
%An example from each of TyDiQA-GoldP and MGSM is shown in Figure \ref{fig:multilingual_examples}. 

In comparing the performance of \ours{} 15B on TyDiQA-GoldP to a range of models, Table \ref{tab:tydiqa_internal} shows that \ours{} 15B achieves the best performance. Impressively, \ours{} 15B is able to significantly improve upon the next best model, PaLM 62B-cont. 

%\shrimai{ add mistral no from sheet}
%Additionally, when we consider the 8-shot results of NeMoTron, we find that it is able to outperform the 1-shot results of PaLM 62B-cont despite the fact that the later has had continued training on top of pretraining.  

\begin{table}[H]
\centering
\begin{adjustbox}{max width=\textwidth, scale=1.0}
\begin{tabular}{crrrrrrrrrrrr}
\toprule
Model & Size & AR & BN & FI & ID & KO & RU & SW & TE & AVG  \\
\midrule
\multirow{2}{*}{PaLM}& 62B & 31.2 & 42.5 & 41.7 & 41.6	& 49.3 & 29.2 &	58.1 & 30.6 &	40.5    \\
 & 62B-cont & 39.4	& 48.7 & 44.0 &	49.2	&52.5&	35.6 &	60.9&	35.3	& 45.7    \\
 LLaMA-2 & 13B & -& -& -& -& -& -& -& -& 33.2 \\
 Baichuan-2 & 13B & -& -& -& -& -& -& -& -& 30.8\\
 QWEN & 14B & -& -& -& -& -& -& -& -& 39.8 \\
 %Mistral & 7B &  -& -& -& -& -& -& -& -& 22.5 \\
\ours{} & 15B & 39.1 & 55.8 &52.2 & 54.5 & 55.1 & 37.8 & 54.5 & 55.0 & \textbf{50.5} \\
%\midrule
%8-Shot & \ours{} & 15B &  \\
\bottomrule
\end{tabular}
\end{adjustbox}
\caption{\label{tab:tydiqa_internal} Comparative results in the one-shot setting on TyDiQA-GoldP. Results for LLaMA-2 13B, Baichuan-2 13B and QWEN 14B are taken from \citep{chen2024orion14b}.  }
\end{table}
%while results for Mistral are from our runs of Mistral in the same setting as \ours{}.

Further demonstrating the impressive multilingual ability of \ours{} 15B, Table \ref{tab:mgsm_all} shows the performance on MGSM. 
We report using the English chain-of-thought setting introduced in \citep{shi2022language} where all chain of thought explanations are presented to the model in English rather than in the language of the task. On this challenging task which assesses the intersection of mathematical and multilingual ability, \ours{} 15B achieves the best performance amongst compared models and improves upon the closest score by nearly 30\%.

% , and strong performance on this benchmark indicates its superior capabilities in these domains.

%However, we see that NeMoTron struggles in comparison to the best results, often with models of multiple hundreds of billions of parameters, that have been reported in literature.  

%\shrimai{Check with Jupinder if avg includes Eng? if 62B-cont nos are available? any other better model? Add Mistral no from sheet -- to add }

\begin{table}[H]
\centering
\begin{adjustbox}{max width=\textwidth, scale=1.0}
\begin{tabular}{lcrrrrrrrrrrrrrrr}
\toprule
Mode & Model & Size & DE & FR & ES & RU & ZH & JA & TH & TE & BN & SW & AVG  \\
\midrule
 \multirow{1}{*}{Native-COT}  & \multirow{1}{*}{PaLM} 
 &   62B 	&24.0  &	24.0 &	26.0 &	22.8 &	24.8	& 14.8 & 18.0 & 11.6 &	13.6 & 9.6&	18.9	\\
\midrule
\multirow{3}{*}{English-COT} & \multirow{1}{*}{PALM} & 62B-cont & 44.8 & 39.2 & 44.4 & 36.8 & 33.6 & 24.0 & 28.0 & 19.6 & 28.0 & 21.2 & 32.0 \\
& \multirow{1}{*}{Mistral} & 7B & 33.2 & 35.2 & 35.6 & 35.2 & 33.2 & 18.8 & 10.0 & 0.0 & 8.0 & 9.2 & 21.8 \\
& \multirow{1}{*}{\ours{}} & 15B & 46.8 & 46.0 & 50.0&45.6&40.0&40.0&43.6&41.6&43.6&16.0 & \textbf{41.3} \\
\bottomrule
\end{tabular}
\end{adjustbox}
\caption{\label{tab:mgsm_all} Eight-shot accuracy results on MGSM. Results for Mistral are from our runs of Mistral in the same setting as \ours{}. }
\end{table}

\textbf{3. Machine Translation:} We additionally evaluate the translation ability of our models through the FLORES-101 \citep{goyal2021} benchmark. The ability to translate between languages is a good test of the model's ability to relate and understand semantic relationships between languages.

As seen in Table~\ref{tab:flores_ext}, \ours{} 15B heftily outperforms both LLaMA-2 13B and Baichuan-2 13B -- improving upon their performance by 90.2\% and 44.1\% respectively. \ours{} 15B does not solely perform well on translating from Chinese into English but is able to attain impressive results on the direct translation of Chinese into other languages. This ability highlights the strong understanding that \ours{} 15B has across a broad spectrum of natural languages.

\begin{table}[H]
\centering
\begin{adjustbox}{max width=\textwidth, scale=1.0}
\begin{tabular}{lcrrrrrrrr}
\toprule
  & Size & ZH-EN & ZH-FR & ZH-ES & ZH-AR & ZH-RU & ZH-JA & ZH-DE & AVG  \\
\midrule
\multirow{1}{*}{LLaMA-2} 
& 13B & 25.4 &19.2 &17.5& 1.4& 10.3& 0.1 &11.1 &12.2  \\
\midrule
\multirow{1}{*}{Baichuan-2} 
& 13B & 30.6 & 22.1 & 17.3 & 2.4 & 14.2 & 11.6 & 14.5 & 16.1 \\
\midrule
\ours{} & 15B & 34.0 & 28.1 & 21.3 & 16.8 & 21.2 & 23.1 &	18.1 & \textbf{23.2} \\ 
\bottomrule
\end{tabular}
\end{adjustbox}
\caption{\label{tab:flores_ext} Eight-shot results on Flores sub-tasks translating out of Chinese. All results for external models were obtained from \citep{yang2023baichuan}}
\end{table}

%In Supplementary Materials \ref{supp:multi}, for each benchmark, we share the used prompt and any additional specific parameters. 

%\section{Instruction Tuning}

\section{Conclusion}

We present \ours{} 15B, a decoder-only transformer-based large language model.
It is trained on 8 trillion tokens spanning English, 53 additional natural languages as well as 43 programming languages. \ours{} 15B exhibits the strongest multilingual performance of any general purpose language model at its scale -- even outperforming models specialized for the multilingual domain.
\ours{} demonstrates that pre-training sets for large language models can continue to be scaled up even further in order to improve the abilities of models. 

%large language models can be pre-trained on more tokens than previously estimated with exceptional results. 

%and that smaller models with more tokens can outperform models more than twice as large.

%\ours{} performs better than specialized multilingual models on XCops, TyDiQA, Flores and MGSM benchmarks and better than code specialized Starcoder model on Multiple-e tasks especially focusing on low-resource languages.

%This opens the possibility of the field exploring the limits of the scaling laws and going beyond.

%\section{Acknowledgements}

%\newpage

\bibliographystyle{plainnat}
\bibliography{references}

\begin{thebibliography}{50}
\providecommand{\natexlab}[1]{#1}
\providecommand{\url}[1]{\texttt{#1}}
\expandafter\ifx\csname urlstyle\endcsname\relax
  \providecommand{\doi}[1]{doi: #1}\else
  \providecommand{\doi}{doi: \begingroup \urlstyle{rm}\Url}\fi

\bibitem[nvl()]{nvlink}
N{VL}ink and {NVS}witch.
\newblock \url{https://www.nvidia.com/en-us/data-center/nvlink/}.

\bibitem[Ainslie et~al.(2023)Ainslie, Lee-Thorp, de~Jong, Zemlyanskiy, Lebr{\'o}n, and Sanghai]{ainslie2023gqa}
Joshua Ainslie, James Lee-Thorp, Michiel de~Jong, Yury Zemlyanskiy, Federico Lebr{\'o}n, and Sumit Sanghai.
\newblock {GQA: Training Generalized Multi-Query Transformer Models from Multi-Head Checkpoints}.
\newblock \emph{arXiv preprint arXiv:2305.13245}, 2023.

\bibitem[Allal et~al.(2023)Allal, Li, Kocetkov, Mou, Akiki, Ferrandis, Muennighoff, Mishra, Gu, Dey, Umapathi, Anderson, Zi, Poirier, Schoelkopf, Troshin, Abulkhanov, Romero, Lappert, Toni, del Río, Liu, Bose, Bhattacharyya, Zhuo, Yu, Villegas, Zocca, Mangrulkar, Lansky, Nguyen, Contractor, Villa, Li, Bahdanau, Jernite, Hughes, Fried, Guha, de~Vries, and von Werra]{allal2023santacoder}
Loubna~Ben Allal, Raymond Li, Denis Kocetkov, Chenghao Mou, Christopher Akiki, Carlos~Munoz Ferrandis, Niklas Muennighoff, Mayank Mishra, Alex Gu, Manan Dey, Logesh~Kumar Umapathi, Carolyn~Jane Anderson, Yangtian Zi, Joel~Lamy Poirier, Hailey Schoelkopf, Sergey Troshin, Dmitry Abulkhanov, Manuel Romero, Michael Lappert, Francesco~De Toni, Bernardo~García del Río, Qian Liu, Shamik Bose, Urvashi Bhattacharyya, Terry~Yue Zhuo, Ian Yu, Paulo Villegas, Marco Zocca, Sourab Mangrulkar, David Lansky, Huu Nguyen, Danish Contractor, Luis Villa, Jia Li, Dzmitry Bahdanau, Yacine Jernite, Sean Hughes, Daniel Fried, Arjun Guha, Harm de~Vries, and Leandro von Werra.
\newblock {SantaCoder: Don't Reach for the Stars!}, 2023.

\bibitem[Austin et~al.(2021)Austin, Odena, Nye, Bosma, Michalewski, Dohan, Jiang, Cai, Terry, Le, and Sutton]{austin2021program}
Jacob Austin, Augustus Odena, Maxwell Nye, Maarten Bosma, Henryk Michalewski, David Dohan, Ellen Jiang, Carrie Cai, Michael Terry, Quoc Le, and Charles Sutton.
\newblock {Program Synthesis with Large Language Models}, 2021.

\bibitem[Bai et~al.(2023)Bai, Bai, Chu, Cui, Dang, Deng, Fan, Ge, Han, Huang, et~al.]{bai2023qwen}
Jinze Bai, Shuai Bai, Yunfei Chu, Zeyu Cui, Kai Dang, Xiaodong Deng, Yang Fan, Wenbin Ge, Yu~Han, Fei Huang, et~al.
\newblock {Qwen Technical Report}.
\newblock \emph{arXiv preprint arXiv:2309.16609}, 2023.

\bibitem[Bisk et~al.(2020)Bisk, Zellers, Bras, Gao, and Choi]{Bisk2020PIQARA}
Yonatan Bisk, Rowan Zellers, Ronan~Le Bras, Jianfeng Gao, and Yejin Choi.
\newblock {PIQA: Reasoning about Physical Commonsense in Natural Language}.
\newblock In \emph{AAAI}, 2020.

\bibitem[Brown et~al.(2020)Brown, Mann, Ryder, Subbiah, Kaplan, Dhariwal, Neelakantan, Shyam, Sastry, Askell, Agarwal, Herbert{-}Voss, Krueger, Henighan, Child, Ramesh, Ziegler, Wu, Winter, Hesse, Chen, Sigler, Litwin, Gray, Chess, Clark, Berner, McCandlish, Radford, Sutskever, and Amodei]{brown2020language}
Tom~B. Brown, Benjamin Mann, Nick Ryder, Melanie Subbiah, Jared Kaplan, Prafulla Dhariwal, Arvind Neelakantan, Pranav Shyam, Girish Sastry, Amanda Askell, Sandhini Agarwal, Ariel Herbert{-}Voss, Gretchen Krueger, Tom Henighan, Rewon Child, Aditya Ramesh, Daniel~M. Ziegler, Jeffrey Wu, Clemens Winter, Christopher Hesse, Mark Chen, Eric Sigler, Mateusz Litwin, Scott Gray, Benjamin Chess, Jack Clark, Christopher Berner, Sam McCandlish, Alec Radford, Ilya Sutskever, and Dario Amodei.
\newblock {Language Models are Few-Shot Learners}.
\newblock In Hugo Larochelle, Marc'Aurelio Ranzato, Raia Hadsell, Maria{-}Florina Balcan, and Hsuan{-}Tien Lin, editors, \emph{Advances in Neural Information Processing Systems 33: Annual Conference on Neural Information Processing Systems 2020, NeurIPS 2020, December 6-12, 2020, virtual}, 2020.
\newblock URL \url{https://proceedings.neurips.cc/paper/2020/hash/1457c0d6bfcb4967418bfb8ac142f64a-Abstract.html}.

\bibitem[Cassano et~al.(2023{\natexlab{a}})Cassano, Gouwar, Nguyen, Nguyen, Phipps-Costin, Pinckney, Yee, Zi, Anderson, Feldman, Guha, Greenberg, and Jangda]{multiple2023}
Federico Cassano, John Gouwar, Daniel Nguyen, Sydney Nguyen, Luna Phipps-Costin, Donald Pinckney, Ming-Ho Yee, Yangtian Zi, Carolyn~Jane Anderson, Molly~Q Feldman, Arjun Guha, Michael Greenberg, and Abhinav Jangda.
\newblock {MultiPL-E: A Scalable and Polyglot Approach to Benchmarking Neural Code Generation}.
\newblock \emph{IEEE Transactions on Software Engineering}, pages 1--17, 2023{\natexlab{a}}.
\newblock \doi{10.1109/TSE.2023.3267446}.

\bibitem[Cassano et~al.(2023{\natexlab{b}})Cassano, Gouwar, Nguyen, Nguyen, Phipps-Costin, Pinckney, Yee, Zi, Anderson, Feldman, et~al.]{cassano2023multipl}
Federico Cassano, John Gouwar, Daniel Nguyen, Sydney Nguyen, Luna Phipps-Costin, Donald Pinckney, Ming-Ho Yee, Yangtian Zi, Carolyn~Jane Anderson, Molly~Q Feldman, et~al.
\newblock Multipl-e: a scalable and polyglot approach to benchmarking neural code generation.
\newblock \emph{IEEE Transactions on Software Engineering}, 2023{\natexlab{b}}.

\bibitem[Chen et~al.(2024)Chen, Huang, Li, Li, Liu, Pan, Xu, Zhang, Zhang, and Han]{chen2024orion14b}
Du~Chen, Yi~Huang, Xiaopu Li, Yongqiang Li, Yongqiang Liu, Haihui Pan, Leichao Xu, Dacheng Zhang, Zhipeng Zhang, and Kun Han.
\newblock {Orion-14B: Open-source Multilingual Large Language Models}, 2024.

\bibitem[Chen et~al.(2021)Chen, Tworek, Jun, Yuan, de~Oliveira~Pinto, Kaplan, Edwards, Burda, Joseph, Brockman, Ray, Puri, Krueger, Petrov, Khlaaf, Sastry, Mishkin, Chan, Gray, Ryder, Pavlov, Power, Kaiser, Bavarian, Winter, Tillet, Such, Cummings, Plappert, Chantzis, Barnes, Herbert-Voss, Guss, Nichol, Paino, Tezak, Tang, Babuschkin, Balaji, Jain, Saunders, Hesse, Carr, Leike, Achiam, Misra, Morikawa, Radford, Knight, Brundage, Murati, Mayer, Welinder, McGrew, Amodei, McCandlish, Sutskever, and Zaremba]{chen2021evaluating}
Mark Chen, Jerry Tworek, Heewoo Jun, Qiming Yuan, Henrique~Ponde de~Oliveira~Pinto, Jared Kaplan, Harri Edwards, Yuri Burda, Nicholas Joseph, Greg Brockman, Alex Ray, Raul Puri, Gretchen Krueger, Michael Petrov, Heidy Khlaaf, Girish Sastry, Pamela Mishkin, Brooke Chan, Scott Gray, Nick Ryder, Mikhail Pavlov, Alethea Power, Lukasz Kaiser, Mohammad Bavarian, Clemens Winter, Philippe Tillet, Felipe~Petroski Such, Dave Cummings, Matthias Plappert, Fotios Chantzis, Elizabeth Barnes, Ariel Herbert-Voss, William~Hebgen Guss, Alex Nichol, Alex Paino, Nikolas Tezak, Jie Tang, Igor Babuschkin, Suchir Balaji, Shantanu Jain, William Saunders, Christopher Hesse, Andrew~N. Carr, Jan Leike, Josh Achiam, Vedant Misra, Evan Morikawa, Alec Radford, Matthew Knight, Miles Brundage, Mira Murati, Katie Mayer, Peter Welinder, Bob McGrew, Dario Amodei, Sam McCandlish, Ilya Sutskever, and Wojciech Zaremba.
\newblock {Evaluating Large Language Models Trained on Code}, 2021.

\bibitem[Chowdhery et~al.(2022)Chowdhery, Narang, Devlin, Bosma, Mishra, Roberts, Barham, Chung, Sutton, Gehrmann, et~al.]{chowdhery2022palm}
Aakanksha Chowdhery, Sharan Narang, Jacob Devlin, Maarten Bosma, Gaurav Mishra, Adam Roberts, Paul Barham, Hyung~Won Chung, Charles Sutton, Sebastian Gehrmann, et~al.
\newblock {PaLM: Scaling Language Modeling with Pathways}.
\newblock \emph{arXiv preprint arXiv:2204.02311}, 2022.

\bibitem[Clark et~al.(2020)Clark, Choi, Collins, Garrette, Kwiatkowski, Nikolaev, and Palomaki]{tydiqa}
Jonathan~H. Clark, Eunsol Choi, Michael Collins, Dan Garrette, Tom Kwiatkowski, Vitaly Nikolaev, and Jennimaria Palomaki.
\newblock {TyDi {QA:} {A} Benchmark for Information-Seeking Question Answering in Typologically Diverse Languages}.
\newblock \emph{CoRR}, abs/2003.05002, 2020.
\newblock URL \url{https://arxiv.org/abs/2003.05002}.

\bibitem[Clark et~al.(2018)Clark, Cowhey, Etzioni, Khot, Sabharwal, Schoenick, and Tafjord]{clark2018think}
Peter Clark, Isaac Cowhey, Oren Etzioni, Tushar Khot, Ashish Sabharwal, Carissa Schoenick, and Oyvind Tafjord.
\newblock {Think You have Solved Question Answering? Try ARC, the AI2 Reasoning Challenge}.
\newblock \emph{arXiv preprint arXiv:1803.05457}, 2018.

\bibitem[Cobbe et~al.(2021)Cobbe, Kosaraju, Bavarian, Hilton, Nakano, Hesse, and Schulman]{cobbe2021}
Karl Cobbe, Vineet Kosaraju, Mohammad Bavarian, Jacob Hilton, Reiichiro Nakano, Christopher Hesse, and John Schulman.
\newblock {Training Verifiers to Solve Math Word Problems}.
\newblock \emph{CoRR}, abs/2110.14168, 2021.
\newblock URL \url{https://arxiv.org/abs/2110.14168}.

\bibitem[Gao et~al.(2021)Gao, Tow, Biderman, Black, DiPofi, Foster, Golding, Hsu, McDonell, Muennighoff, Phang, Reynolds, Tang, Thite, Wang, Wang, and Zou]{eval-harness}
Leo Gao, Jonathan Tow, Stella Biderman, Sid Black, Anthony DiPofi, Charles Foster, Laurence Golding, Jeffrey Hsu, Kyle McDonell, Niklas Muennighoff, Jason Phang, Laria Reynolds, Eric Tang, Anish Thite, Ben Wang, Kevin Wang, and Andy Zou.
\newblock {A Framework for Few-shot Language Model Evaluation}, September 2021.
\newblock URL \url{https://doi.org/10.5281/zenodo.5371628}.

\bibitem[Gemma~Team(2024)]{gemma24}
Google~DeepMind Gemma~Team.
\newblock {Gemma: Open Models Based on Gemini Research and Technology}, 2024.

\bibitem[Google(2023)]{geminiteam2023gemini}
Google.
\newblock {Gemini: A Family of Highly Capable Multimodal Models}, 2023.

\bibitem[Goyal et~al.(2021)Goyal, Gao, Chaudhary, Chen, Wenzek, Ju, Krishnan, Ranzato, Guzm{\'{a}}n, and Fan]{goyal2021}
Naman Goyal, Cynthia Gao, Vishrav Chaudhary, Peng{-}Jen Chen, Guillaume Wenzek, Da~Ju, Sanjana Krishnan, Marc'Aurelio Ranzato, Francisco Guzm{\'{a}}n, and Angela Fan.
\newblock {The {FLORES-101} Evaluation Benchmark for Low-Resource and Multilingual Machine Translation}.
\newblock \emph{CoRR}, abs/2106.03193, 2021.
\newblock URL \url{https://arxiv.org/abs/2106.03193}.

\bibitem[Hendrycks et~al.(2020)Hendrycks, Burns, Basart, Zou, Mazeika, Song, and Steinhardt]{hendrycks2020measuring}
Dan Hendrycks, Collin Burns, Steven Basart, Andy Zou, Mantas Mazeika, Dawn Song, and Jacob Steinhardt.
\newblock {Measuring Massive Multitask Language Understanding}.
\newblock \emph{arXiv preprint arXiv:2009.03300}, 2020.

\bibitem[Hoffmann et~al.(2022)Hoffmann, Borgeaud, Mensch, Buchatskaya, Cai, Rutherford, Casas, Hendricks, Welbl, Clark, et~al.]{hoffmann2022training}
Jordan Hoffmann, Sebastian Borgeaud, Arthur Mensch, Elena Buchatskaya, Trevor Cai, Eliza Rutherford, Diego de~Las Casas, Lisa~Anne Hendricks, Johannes Welbl, Aidan Clark, et~al.
\newblock {Training Compute-Optimal Large Language Models}.
\newblock \emph{arXiv preprint arXiv:2203.15556}, 2022.

\bibitem[Jennings et~al.(2023)Jennings, Patwary, Subramanian, Prabhumoye, Dattagupta, Shoeybi, and Catanzaro]{jennings2023ndc}
Joseph Jennings, Mostofa Patwary, Sandeep Subramanian, Shrimai Prabhumoye, Ayush Dattagupta, Mohammad Shoeybi, and Bryan Catanzaro.
\newblock {Curating Trillion-Token Datasets: Introducing NVIDIA NeMo Data Curator}.
\newblock \url{https://developer.nvidia.com/blog/curating-trillion-token-datasets-introducing-nemo-data-curator/}, 2023.

\bibitem[Jiang et~al.(2023)Jiang, Sablayrolles, Mensch, Bamford, Chaplot, Casas, Bressand, Lengyel, Lample, Saulnier, et~al.]{jiang2023mistral}
Albert~Q Jiang, Alexandre Sablayrolles, Arthur Mensch, Chris Bamford, Devendra~Singh Chaplot, Diego de~las Casas, Florian Bressand, Gianna Lengyel, Guillaume Lample, Lucile Saulnier, et~al.
\newblock {Mistral 7B}.
\newblock \emph{arXiv preprint arXiv:2310.06825}, 2023.

\bibitem[Kaplan et~al.(2020)Kaplan, McCandlish, Henighan, Brown, Chess, Child, Gray, Radford, Wu, and Amodei]{kaplan2020scaling}
Jared Kaplan, Sam McCandlish, Tom Henighan, Tom~B Brown, Benjamin Chess, Rewon Child, Scott Gray, Alec Radford, Jeffrey Wu, and Dario Amodei.
\newblock {Scaling Laws for Neural Language Models}.
\newblock \emph{arXiv preprint arXiv:2001.08361}, 2020.

\bibitem[Korthikanti et~al.(2022)Korthikanti, Casper, Lym, McAfee, Andersch, Shoeybi, and Catanzaro]{korthikanti2022sequence}
Vijay Korthikanti, Jared Casper, Sangkug Lym, Lawrence McAfee, Michael Andersch, Mohammad Shoeybi, and Bryan Catanzaro.
\newblock {Reducing Activation Recomputation in Large Transformer Models}, 2022.

\bibitem[Kudo and Richardson(2018)]{kudo2018sentencepiece}
Taku Kudo and John Richardson.
\newblock {Sentencepiece: A Simple and Language Independent Subword Tokenizer and Detokenizer for Neural Text Processing}.
\newblock \emph{arXiv preprint arXiv:1808.06226}, 2018.

\bibitem[Li et~al.(2023)Li, Allal, Zi, Muennighoff, Kocetkov, Mou, Marone, Akiki, Li, Chim, Liu, Zheltonozhskii, Zhuo, Wang, Dehaene, Davaadorj, Lamy-Poirier, Monteiro, Shliazhko, Gontier, Meade, Zebaze, Yee, Umapathi, Zhu, Lipkin, Oblokulov, Wang, Murthy, Stillerman, Patel, Abulkhanov, Zocca, Dey, Zhang, Fahmy, Bhattacharyya, Yu, Singh, Luccioni, Villegas, Kunakov, Zhdanov, Romero, Lee, Timor, Ding, Schlesinger, Schoelkopf, Ebert, Dao, Mishra, Gu, Robinson, Anderson, Dolan-Gavitt, Contractor, Reddy, Fried, Bahdanau, Jernite, Ferrandis, Hughes, Wolf, Guha, von Werra, and de~Vries]{li2023starcoder}
Raymond Li, Loubna~Ben Allal, Yangtian Zi, Niklas Muennighoff, Denis Kocetkov, Chenghao Mou, Marc Marone, Christopher Akiki, Jia Li, Jenny Chim, Qian Liu, Evgenii Zheltonozhskii, Terry~Yue Zhuo, Thomas Wang, Olivier Dehaene, Mishig Davaadorj, Joel Lamy-Poirier, João Monteiro, Oleh Shliazhko, Nicolas Gontier, Nicholas Meade, Armel Zebaze, Ming-Ho Yee, Logesh~Kumar Umapathi, Jian Zhu, Benjamin Lipkin, Muhtasham Oblokulov, Zhiruo Wang, Rudra Murthy, Jason Stillerman, Siva~Sankalp Patel, Dmitry Abulkhanov, Marco Zocca, Manan Dey, Zhihan Zhang, Nour Fahmy, Urvashi Bhattacharyya, Wenhao Yu, Swayam Singh, Sasha Luccioni, Paulo Villegas, Maxim Kunakov, Fedor Zhdanov, Manuel Romero, Tony Lee, Nadav Timor, Jennifer Ding, Claire Schlesinger, Hailey Schoelkopf, Jan Ebert, Tri Dao, Mayank Mishra, Alex Gu, Jennifer Robinson, Carolyn~Jane Anderson, Brendan Dolan-Gavitt, Danish Contractor, Siva Reddy, Daniel Fried, Dzmitry Bahdanau, Yacine Jernite, Carlos~Muñoz Ferrandis, Sean Hughes, Thomas Wolf, Arjun Guha, Leandro von
  Werra, and Harm de~Vries.
\newblock {StarCoder: May the Source be with You!}, 2023.

\bibitem[Lin et~al.(2022)Lin, Mihaylov, Artetxe, Wang, Chen, Simig, Ott, Goyal, Bhosale, Du, Pasunuru, Shleifer, Koura, Chaudhary, O'Horo, Wang, Zettlemoyer, Kozareva, Diab, Stoyanov, and Li]{lin2022fewshot}
Xi~Victoria Lin, Todor Mihaylov, Mikel Artetxe, Tianlu Wang, Shuohui Chen, Daniel Simig, Myle Ott, Naman Goyal, Shruti Bhosale, Jingfei Du, Ramakanth Pasunuru, Sam Shleifer, Punit~Singh Koura, Vishrav Chaudhary, Brian O'Horo, Jeff Wang, Luke Zettlemoyer, Zornitsa Kozareva, Mona Diab, Veselin Stoyanov, and Xian Li.
\newblock {Few-shot Learning with Multilingual Language Models}, 2022.

\bibitem[NVIDIA(2022)]{nvidia2022h100}
NVIDIA.
\newblock {H100 Tensor Core GPU Architecture Overview}, 2022.

\bibitem[Papineni et~al.(2002)Papineni, Roukos, Ward, and Zhu]{papineni2002}
Kishore Papineni, Salim Roukos, Todd Ward, and Wei-Jing Zhu.
\newblock {BLEU: A Method for Automatic Evaluation of Machine Translation}.
\newblock In \emph{Proceedings of the 40th Annual Meeting on Association for Computational Linguistics}, ACL '02, page 311–318, USA, 2002. Association for Computational Linguistics.
\newblock \doi{10.3115/1073083.1073135}.
\newblock URL \url{https://doi.org/10.3115/1073083.1073135}.

\bibitem[Ponti et~al.(2020)Ponti, Glavas, Majewska, Liu, Vulic, and Korhonen]{xcopa}
Edoardo~Maria Ponti, Goran Glavas, Olga Majewska, Qianchu Liu, Ivan Vulic, and Anna Korhonen.
\newblock {{XCOPA:} {A} Multilingual Dataset for Causal Commonsense Reasoning}.
\newblock \emph{CoRR}, abs/2005.00333, 2020.
\newblock URL \url{https://arxiv.org/abs/2005.00333}.

\bibitem[Post(2018)]{post2018}
Matt Post.
\newblock {A Call for Clarity in Reporting {BLEU} Scores}.
\newblock \emph{CoRR}, abs/1804.08771, 2018.
\newblock URL \url{http://arxiv.org/abs/1804.08771}.

\bibitem[Rae et~al.(2022)Rae, Borgeaud, Cai, Millican, Hoffmann, Song, Aslanides, Henderson, Ring, Young, Rutherford, Hennigan, Menick, Cassirer, Powell, van~den Driessche, Hendricks, Rauh, Huang, Glaese, Welbl, Dathathri, Huang, Uesato, Mellor, Higgins, Creswell, McAleese, Wu, Elsen, Jayakumar, Buchatskaya, Budden, Sutherland, Simonyan, Paganini, Sifre, Martens, Li, Kuncoro, Nematzadeh, Gribovskaya, Donato, Lazaridou, Mensch, Lespiau, Tsimpoukelli, Grigorev, Fritz, Sottiaux, Pajarskas, Pohlen, Gong, Toyama, de~Masson~d'Autume, Li, Terzi, Mikulik, Babuschkin, Clark, de~Las~Casas, Guy, Jones, Bradbury, Johnson, Hechtman, Weidinger, Gabriel, Isaac, Lockhart, Osindero, Rimell, Dyer, Vinyals, Ayoub, Stanway, Bennett, Hassabis, Kavukcuoglu, and Irving]{rae2022scaling}
Jack~W. Rae, Sebastian Borgeaud, Trevor Cai, Katie Millican, Jordan Hoffmann, Francis Song, John Aslanides, Sarah Henderson, Roman Ring, Susannah Young, Eliza Rutherford, Tom Hennigan, Jacob Menick, Albin Cassirer, Richard Powell, George van~den Driessche, Lisa~Anne Hendricks, Maribeth Rauh, Po-Sen Huang, Amelia Glaese, Johannes Welbl, Sumanth Dathathri, Saffron Huang, Jonathan Uesato, John Mellor, Irina Higgins, Antonia Creswell, Nat McAleese, Amy Wu, Erich Elsen, Siddhant Jayakumar, Elena Buchatskaya, David Budden, Esme Sutherland, Karen Simonyan, Michela Paganini, Laurent Sifre, Lena Martens, Xiang~Lorraine Li, Adhiguna Kuncoro, Aida Nematzadeh, Elena Gribovskaya, Domenic Donato, Angeliki Lazaridou, Arthur Mensch, Jean-Baptiste Lespiau, Maria Tsimpoukelli, Nikolai Grigorev, Doug Fritz, Thibault Sottiaux, Mantas Pajarskas, Toby Pohlen, Zhitao Gong, Daniel Toyama, Cyprien de~Masson~d'Autume, Yujia Li, Tayfun Terzi, Vladimir Mikulik, Igor Babuschkin, Aidan Clark, Diego de~Las~Casas, Aurelia Guy, Chris Jones,
  James Bradbury, Matthew Johnson, Blake Hechtman, Laura Weidinger, Iason Gabriel, William Isaac, Ed~Lockhart, Simon Osindero, Laura Rimell, Chris Dyer, Oriol Vinyals, Kareem Ayoub, Jeff Stanway, Lorrayne Bennett, Demis Hassabis, Koray Kavukcuoglu, and Geoffrey Irving.
\newblock {Scaling Language Models: Methods, Analysis \& Insights from Training Gopher}, 2022.

\bibitem[Raffel et~al.(2020)Raffel, Shazeer, Roberts, Lee, Narang, Matena, Zhou, Li, and Liu]{raffel2020exploring}
Colin Raffel, Noam Shazeer, Adam Roberts, Katherine Lee, Sharan Narang, Michael Matena, Yanqi Zhou, Wei Li, and Peter~J Liu.
\newblock {Exploring the Limits of Transfer Learning with a Unified Text-to-Text Transformer}.
\newblock \emph{The Journal of Machine Learning Research}, 21\penalty0 (1):\penalty0 5485--5551, 2020.

\bibitem[Sakaguchi et~al.(2020)Sakaguchi, Bras, Bhagavatula, and Choi]{Sakaguchi2020WINOGRANDEAA}
Keisuke Sakaguchi, Ronan~Le Bras, Chandra Bhagavatula, and Yejin Choi.
\newblock {WINOGRANDE: An Adversarial Winograd Schema Challenge at Scale}.
\newblock In \emph{AAAI}, 2020.

\bibitem[Sap et~al.(2019)Sap, Rashkin, Chen, LeBras, and Choi]{sap2019socialiqa}
Maarten Sap, Hannah Rashkin, Derek Chen, Ronan LeBras, and Yejin Choi.
\newblock Socialiqa: Commonsense reasoning about social interactions, 2019.

\bibitem[Scao et~al.(2023)Scao, Fan, Akiki, Pavlick, Ilić, Hesslow, Castagné, Luccioni, Yvon, Gallé, Tow, Rush, Biderman, Webson, Ammanamanchi, Wang, Sagot, Muennighoff, del Moral, Ruwase, Bawden, Bekman, McMillan-Major, Beltagy, Nguyen, Saulnier, Tan, Suarez, Sanh, Laurençon, Jernite, Launay, Mitchell, Raffel, Gokaslan, Simhi, Soroa, Aji, Alfassy, Rogers, Nitzav, Xu, Mou, Emezue, Klamm, Leong, van Strien, Adelani, Radev, Ponferrada, Levkovizh, Kim, Natan, Toni, Dupont, Kruszewski, Pistilli, Elsahar, Benyamina, Tran, Yu, Abdulmumin, Johnson, Gonzalez-Dios, de~la Rosa, Chim, Dodge, Zhu, Chang, Frohberg, Tobing, Bhattacharjee, Almubarak, Chen, Lo, Werra, Weber, Phan, allal, Tanguy, Dey, Muñoz, Masoud, Grandury, Šaško, Huang, Coavoux, Singh, Jiang, Vu, Jauhar, Ghaleb, Subramani, Kassner, Khamis, Nguyen, Espejel, de~Gibert, Villegas, Henderson, Colombo, Amuok, Lhoest, Harliman, Bommasani, López, Ribeiro, Osei, Pyysalo, Nagel, Bose, Muhammad, Sharma, Longpre, Nikpoor, Silberberg, Pai, Zink, Torrent,
  Schick, Thrush, Danchev, Nikoulina, Laippala, Lepercq, Prabhu, Alyafeai, Talat, Raja, Heinzerling, Si, Taşar, Salesky, Mielke, Lee, Sharma, Santilli, Chaffin, Stiegler, Datta, Szczechla, Chhablani, Wang, Pandey, Strobelt, Fries, Rozen, Gao, Sutawika, Bari, Al-shaibani, Manica, Nayak, Teehan, Albanie, Shen, Ben-David, Bach, Kim, Bers, Fevry, Neeraj, Thakker, Raunak, Tang, Yong, Sun, Brody, Uri, Tojarieh, Roberts, Chung, Tae, Phang, Press, Li, Narayanan, Bourfoune, Casper, Rasley, Ryabinin, Mishra, Zhang, Shoeybi, Peyrounette, Patry, Tazi, Sanseviero, von Platen, Cornette, Lavallée, Lacroix, Rajbhandari, Gandhi, Smith, Requena, Patil, Dettmers, Baruwa, Singh, Cheveleva, Ligozat, Subramonian, Névéol, Lovering, Garrette, Tunuguntla, Reiter, Taktasheva, Voloshina, Bogdanov, Winata, Schoelkopf, Kalo, Novikova, Forde, Clive, Kasai, Kawamura, Hazan, Carpuat, Clinciu, Kim, Cheng, Serikov, Antverg, van~der Wal, Zhang, Zhang, Gehrmann, Mirkin, Pais, Shavrina, Scialom, Yun, Limisiewicz, Rieser, Protasov, Mikhailov,
  Pruksachatkun, Belinkov, Bamberger, Kasner, Rueda, Pestana, Feizpour, Khan, Faranak, Santos, Hevia, Unldreaj, Aghagol, Abdollahi, Tammour, HajiHosseini, Behroozi, Ajibade, Saxena, Ferrandis, Contractor, Lansky, David, Kiela, Nguyen, Tan, Baylor, Ozoani, Mirza, Ononiwu, Rezanejad, Jones, Bhattacharya, Solaiman, Sedenko, Nejadgholi, Passmore, Seltzer, Sanz, Dutra, Samagaio, Elbadri, Mieskes, Gerchick, Akinlolu, McKenna, Qiu, Ghauri, Burynok, Abrar, Rajani, Elkott, Fahmy, Samuel, An, Kromann, Hao, Alizadeh, Shubber, Wang, Roy, Viguier, Le, Oyebade, Le, Yang, Nguyen, Kashyap, Palasciano, Callahan, Shukla, Miranda-Escalada, Singh, Beilharz, Wang, Brito, Zhou, Jain, Xu, Fourrier, Periñán, Molano, Yu, Manjavacas, Barth, Fuhrimann, Altay, Bayrak, Burns, Vrabec, Bello, Dash, Kang, Giorgi, Golde, Posada, Sivaraman, Bulchandani, Liu, Shinzato, de~Bykhovetz, Takeuchi, Pàmies, Castillo, Nezhurina, Sänger, Samwald, Cullan, Weinberg, Wolf, Mihaljcic, Liu, Freidank, Kang, Seelam, Dahlberg, Broad, Muellner, Fung,
  Haller, Chandrasekhar, Eisenberg, Martin, Canalli, Su, Su, Cahyawijaya, Garda, Deshmukh, Mishra, Kiblawi, Ott, Sang-aroonsiri, Kumar, Schweter, Bharati, Laud, Gigant, Kainuma, Kusa, Labrak, Bajaj, Venkatraman, Xu, Xu, Xu, Tan, Xie, Ye, Bras, Belkada, and Wolf]{workshop2023bloom}
Teven~Le Scao, Angela Fan, Christopher Akiki, Ellie Pavlick, Suzana Ilić, Daniel Hesslow, Roman Castagné, Alexandra~Sasha Luccioni, François Yvon, Matthias Gallé, Jonathan Tow, Alexander~M. Rush, Stella Biderman, Albert Webson, Pawan~Sasanka Ammanamanchi, Thomas Wang, Benoît Sagot, Niklas Muennighoff, Albert~Villanova del Moral, Olatunji Ruwase, Rachel Bawden, Stas Bekman, Angelina McMillan-Major, Iz~Beltagy, Huu Nguyen, Lucile Saulnier, Samson Tan, Pedro~Ortiz Suarez, Victor Sanh, Hugo Laurençon, Yacine Jernite, Julien Launay, Margaret Mitchell, Colin Raffel, Aaron Gokaslan, Adi Simhi, Aitor Soroa, Alham~Fikri Aji, Amit Alfassy, Anna Rogers, Ariel~Kreisberg Nitzav, Canwen Xu, Chenghao Mou, Chris Emezue, Christopher Klamm, Colin Leong, Daniel van Strien, David~Ifeoluwa Adelani, Dragomir Radev, Eduardo~González Ponferrada, Efrat Levkovizh, Ethan Kim, Eyal~Bar Natan, Francesco~De Toni, Gérard Dupont, Germán Kruszewski, Giada Pistilli, Hady Elsahar, Hamza Benyamina, Hieu Tran, Ian Yu, Idris Abdulmumin,
  Isaac Johnson, Itziar Gonzalez-Dios, Javier de~la Rosa, Jenny Chim, Jesse Dodge, Jian Zhu, Jonathan Chang, Jörg Frohberg, Joseph Tobing, Joydeep Bhattacharjee, Khalid Almubarak, Kimbo Chen, Kyle Lo, Leandro~Von Werra, Leon Weber, Long Phan, Loubna~Ben allal, Ludovic Tanguy, Manan Dey, Manuel~Romero Muñoz, Maraim Masoud, María Grandury, Mario Šaško, Max Huang, Maximin Coavoux, Mayank Singh, Mike Tian-Jian Jiang, Minh~Chien Vu, Mohammad~A. Jauhar, Mustafa Ghaleb, Nishant Subramani, Nora Kassner, Nurulaqilla Khamis, Olivier Nguyen, Omar Espejel, Ona de~Gibert, Paulo Villegas, Peter Henderson, Pierre Colombo, Priscilla Amuok, Quentin Lhoest, Rheza Harliman, Rishi Bommasani, Roberto~Luis López, Rui Ribeiro, Salomey Osei, Sampo Pyysalo, Sebastian Nagel, Shamik Bose, Shamsuddeen~Hassan Muhammad, Shanya Sharma, Shayne Longpre, Somaieh Nikpoor, Stanislav Silberberg, Suhas Pai, Sydney Zink, Tiago~Timponi Torrent, Timo Schick, Tristan Thrush, Valentin Danchev, Vassilina Nikoulina, Veronika Laippala, Violette
  Lepercq, Vrinda Prabhu, Zaid Alyafeai, Zeerak Talat, Arun Raja, Benjamin Heinzerling, Chenglei Si, Davut~Emre Taşar, Elizabeth Salesky, Sabrina~J. Mielke, Wilson~Y. Lee, Abheesht Sharma, Andrea Santilli, Antoine Chaffin, Arnaud Stiegler, Debajyoti Datta, Eliza Szczechla, Gunjan Chhablani, Han Wang, Harshit Pandey, Hendrik Strobelt, Jason~Alan Fries, Jos Rozen, Leo Gao, Lintang Sutawika, M~Saiful Bari, Maged~S. Al-shaibani, Matteo Manica, Nihal Nayak, Ryan Teehan, Samuel Albanie, Sheng Shen, Srulik Ben-David, Stephen~H. Bach, Taewoon Kim, Tali Bers, Thibault Fevry, Trishala Neeraj, Urmish Thakker, Vikas Raunak, Xiangru Tang, Zheng-Xin Yong, Zhiqing Sun, Shaked Brody, Yallow Uri, Hadar Tojarieh, Adam Roberts, Hyung~Won Chung, Jaesung Tae, Jason Phang, Ofir Press, Conglong Li, Deepak Narayanan, Hatim Bourfoune, Jared Casper, Jeff Rasley, Max Ryabinin, Mayank Mishra, Minjia Zhang, Mohammad Shoeybi, Myriam Peyrounette, Nicolas Patry, Nouamane Tazi, Omar Sanseviero, Patrick von Platen, Pierre Cornette,
  Pierre~François Lavallée, Rémi Lacroix, Samyam Rajbhandari, Sanchit Gandhi, Shaden Smith, Stéphane Requena, Suraj Patil, Tim Dettmers, Ahmed Baruwa, Amanpreet Singh, Anastasia Cheveleva, Anne-Laure Ligozat, Arjun Subramonian, Aurélie Névéol, Charles Lovering, Dan Garrette, Deepak Tunuguntla, Ehud Reiter, Ekaterina Taktasheva, Ekaterina Voloshina, Eli Bogdanov, Genta~Indra Winata, Hailey Schoelkopf, Jan-Christoph Kalo, Jekaterina Novikova, Jessica~Zosa Forde, Jordan Clive, Jungo Kasai, Ken Kawamura, Liam Hazan, Marine Carpuat, Miruna Clinciu, Najoung Kim, Newton Cheng, Oleg Serikov, Omer Antverg, Oskar van~der Wal, Rui Zhang, Ruochen Zhang, Sebastian Gehrmann, Shachar Mirkin, Shani Pais, Tatiana Shavrina, Thomas Scialom, Tian Yun, Tomasz Limisiewicz, Verena Rieser, Vitaly Protasov, Vladislav Mikhailov, Yada Pruksachatkun, Yonatan Belinkov, Zachary Bamberger, Zdeněk Kasner, Alice Rueda, Amanda Pestana, Amir Feizpour, Ammar Khan, Amy Faranak, Ana Santos, Anthony Hevia, Antigona Unldreaj, Arash Aghagol,
  Arezoo Abdollahi, Aycha Tammour, Azadeh HajiHosseini, Bahareh Behroozi, Benjamin Ajibade, Bharat Saxena, Carlos~Muñoz Ferrandis, Danish Contractor, David Lansky, Davis David, Douwe Kiela, Duong~A. Nguyen, Edward Tan, Emi Baylor, Ezinwanne Ozoani, Fatima Mirza, Frankline Ononiwu, Habib Rezanejad, Hessie Jones, Indrani Bhattacharya, Irene Solaiman, Irina Sedenko, Isar Nejadgholi, Jesse Passmore, Josh Seltzer, Julio~Bonis Sanz, Livia Dutra, Mairon Samagaio, Maraim Elbadri, Margot Mieskes, Marissa Gerchick, Martha Akinlolu, Michael McKenna, Mike Qiu, Muhammed Ghauri, Mykola Burynok, Nafis Abrar, Nazneen Rajani, Nour Elkott, Nour Fahmy, Olanrewaju Samuel, Ran An, Rasmus Kromann, Ryan Hao, Samira Alizadeh, Sarmad Shubber, Silas Wang, Sourav Roy, Sylvain Viguier, Thanh Le, Tobi Oyebade, Trieu Le, Yoyo Yang, Zach Nguyen, Abhinav~Ramesh Kashyap, Alfredo Palasciano, Alison Callahan, Anima Shukla, Antonio Miranda-Escalada, Ayush Singh, Benjamin Beilharz, Bo~Wang, Caio Brito, Chenxi Zhou, Chirag Jain, Chuxin Xu,
  Clémentine Fourrier, Daniel~León Periñán, Daniel Molano, Dian Yu, Enrique Manjavacas, Fabio Barth, Florian Fuhrimann, Gabriel Altay, Giyaseddin Bayrak, Gully Burns, Helena~U. Vrabec, Imane Bello, Ishani Dash, Jihyun Kang, John Giorgi, Jonas Golde, Jose~David Posada, Karthik~Rangasai Sivaraman, Lokesh Bulchandani, Lu~Liu, Luisa Shinzato, Madeleine~Hahn de~Bykhovetz, Maiko Takeuchi, Marc Pàmies, Maria~A Castillo, Marianna Nezhurina, Mario Sänger, Matthias Samwald, Michael Cullan, Michael Weinberg, Michiel~De Wolf, Mina Mihaljcic, Minna Liu, Moritz Freidank, Myungsun Kang, Natasha Seelam, Nathan Dahlberg, Nicholas~Michio Broad, Nikolaus Muellner, Pascale Fung, Patrick Haller, Ramya Chandrasekhar, Renata Eisenberg, Robert Martin, Rodrigo Canalli, Rosaline Su, Ruisi Su, Samuel Cahyawijaya, Samuele Garda, Shlok~S Deshmukh, Shubhanshu Mishra, Sid Kiblawi, Simon Ott, Sinee Sang-aroonsiri, Srishti Kumar, Stefan Schweter, Sushil Bharati, Tanmay Laud, Théo Gigant, Tomoya Kainuma, Wojciech Kusa, Yanis Labrak,
  Yash~Shailesh Bajaj, Yash Venkatraman, Yifan Xu, Yingxin Xu, Yu~Xu, Zhe Tan, Zhongli Xie, Zifan Ye, Mathilde Bras, Younes Belkada, and Thomas Wolf.
\newblock {BLOOM: A 176B-Parameter Open-Access Multilingual Language Model}, 2023.

\bibitem[Shi et~al.(2022)Shi, Suzgun, Freitag, Wang, Srivats, Vosoughi, Chung, Tay, Ruder, Zhou, Das, and Wei]{shi2022language}
Freda Shi, Mirac Suzgun, Markus Freitag, Xuezhi Wang, Suraj Srivats, Soroush Vosoughi, Hyung~Won Chung, Yi~Tay, Sebastian Ruder, Denny Zhou, Dipanjan Das, and Jason Wei.
\newblock {Language Models are Multilingual Chain-of-Thought Reasoners}, 2022.

\bibitem[Shliazhko et~al.(2022)Shliazhko, Fenogenova, Tikhonova, Mikhailov, Kozlova, and Shavrina]{shliazhko2022mgpt}
Oleh Shliazhko, Alena Fenogenova, Maria Tikhonova, Vladislav Mikhailov, Anastasia Kozlova, and Tatiana Shavrina.
\newblock {mGPT: Few-Shot Learners Go Multilingual}, 2022.

\bibitem[Shoeybi et~al.(2019)Shoeybi, Patwary, Puri, LeGresley, Casper, and Catanzaro]{shoeybi2019megatron}
Mohammad Shoeybi, Mostofa Patwary, Raul Puri, Patrick LeGresley, Jared Casper, and Bryan Catanzaro.
\newblock {Megatron-LM: Training Multi-Billion Parameter Language Models using Model Parallelism}.
\newblock \emph{arXiv preprint arXiv:1909.08053}, 2019.

\bibitem[Slav~Petrov and et~al.(2023)]{palm2}
Andrew M.~Dai Slav~Petrov, Yonghui~Wu and et~al.
\newblock {PaLM 2 Technical Report}, 2023.
\newblock URL \url{https://ai.google/static/documents/palm2techreport.pdf}.

\bibitem[Smith et~al.(2022)Smith, Patwary, Norick, LeGresley, Rajbhandari, Casper, Liu, Prabhumoye, Zerveas, Korthikanti, Zheng, Child, Aminabadi, Bernauer, Song, Shoeybi, He, Houston, Tiwary, and Catanzaro]{smith2022}
Shaden Smith, Mostofa Patwary, Brandon Norick, Patrick LeGresley, Samyam Rajbhandari, Jared Casper, Zhun Liu, Shrimai Prabhumoye, George Zerveas, Vijay Korthikanti, Elton Zheng, Rewon Child, Reza~Yazdani Aminabadi, Julie Bernauer, Xia Song, Mohammad Shoeybi, Yuxiong He, Michael Houston, Saurabh Tiwary, and Bryan Catanzaro.
\newblock {Using DeepSpeed and Megatron to Train Megatron-Turing {NLG} 530B, {A} Large-Scale Generative Language Model}.
\newblock \emph{CoRR}, abs/2201.11990, 2022.
\newblock URL \url{https://arxiv.org/abs/2201.11990}.

\bibitem[Su et~al.(2021)Su, Lu, Pan, Murtadha, Wen, and Liu]{su2021roformer}
Jianlin Su, Yu~Lu, Shengfeng Pan, Ahmed Murtadha, Bo~Wen, and Yunfeng Liu.
\newblock {Roformer: Enhanced Transformer with Rotary Position Embedding}.
\newblock \emph{arXiv preprint arXiv:2104.09864}, 2021.

\bibitem[Suzgun et~al.(2022)Suzgun, Scales, Schärli, Gehrmann, Tay, Chung, Chowdhery, Le, Chi, Zhou, and Wei]{suzgun2022challenging}
Mirac Suzgun, Nathan Scales, Nathanael Schärli, Sebastian Gehrmann, Yi~Tay, Hyung~Won Chung, Aakanksha Chowdhery, Quoc~V. Le, Ed~H. Chi, Denny Zhou, and Jason Wei.
\newblock Challenging big-bench tasks and whether chain-of-thought can solve them, 2022.

\bibitem[Touvron et~al.(2023{\natexlab{a}})Touvron, Lavril, Izacard, Martinet, Lachaux, Lacroix, Rozière, Goyal, Hambro, Azhar, Rodriguez, Joulin, Grave, and Lample]{touvron2023llama}
Hugo Touvron, Thibaut Lavril, Gautier Izacard, Xavier Martinet, Marie-Anne Lachaux, Timothée Lacroix, Baptiste Rozière, Naman Goyal, Eric Hambro, Faisal Azhar, Aurelien Rodriguez, Armand Joulin, Edouard Grave, and Guillaume Lample.
\newblock {LLaMA: Open and Efficient Foundation Language Models}, 2023{\natexlab{a}}.

\bibitem[Touvron et~al.(2023{\natexlab{b}})Touvron, Martin, Stone, Albert, Almahairi, Babaei, Bashlykov, Batra, Bhargava, Bhosale, et~al.]{touvron2023llama2}
Hugo Touvron, Louis Martin, Kevin Stone, Peter Albert, Amjad Almahairi, Yasmine Babaei, Nikolay Bashlykov, Soumya Batra, Prajjwal Bhargava, Shruti Bhosale, et~al.
\newblock {Llama 2: Open Foundation and Fine-tuned Chat Models}.
\newblock \emph{arXiv preprint arXiv:2307.09288}, 2023{\natexlab{b}}.

\bibitem[Vaswani et~al.(2017)Vaswani, Shazeer, Parmar, Uszkoreit, Jones, Gomez, Kaiser, and Polosukhin]{DBLP:journals/corr/VaswaniSPUJGKP17}
Ashish Vaswani, Noam Shazeer, Niki Parmar, Jakob Uszkoreit, Llion Jones, Aidan~N. Gomez, Lukasz Kaiser, and Illia Polosukhin.
\newblock Attention is all you need.
\newblock \emph{CoRR}, abs/1706.03762, 2017.
\newblock URL \url{http://arxiv.org/abs/1706.03762}.

\bibitem[Wenzek et~al.(2019)Wenzek, Lachaux, Conneau, Chaudhary, Guzm{\'a}n, Joulin, and Grave]{wenzek2019ccnet}
Guillaume Wenzek, Marie-Anne Lachaux, Alexis Conneau, Vishrav Chaudhary, Francisco Guzm{\'a}n, Armand Joulin, and Edouard Grave.
\newblock {CCNet: Extracting High Quality Monolingual Datasets from Web Crawl Data}.
\newblock \emph{arXiv preprint arXiv:1911.00359}, 2019.

\bibitem[Yang et~al.(2023)Yang, Xiao, Wang, Zhang, Yin, Lv, Pan, Wang, Yan, Yang, et~al.]{yang2023baichuan}
Aiyuan Yang, Bin Xiao, Bingning Wang, Borong Zhang, Chao Yin, Chenxu Lv, Da~Pan, Dian Wang, Dong Yan, Fan Yang, et~al.
\newblock {Baichuan 2: Open Large-scale Language Models}.
\newblock \emph{arXiv preprint arXiv:2309.10305}, 2023.

\bibitem[Zellers et~al.(2019)Zellers, Holtzman, Bisk, Farhadi, and Choi]{Zellers2019HellaSwagCA}
Rowan Zellers, Ari Holtzman, Yonatan Bisk, Ali Farhadi, and Yejin Choi.
\newblock {HellaSwag: Can a Machine Really Finish Your Sentence?}
\newblock In \emph{ACL}, 2019.

\end{thebibliography}

\FloatBarrier

\appendix

\section*{Supplementary Materials} \label{supplementary_material}

\begin{table}[H]
\centering
\begin{tabular}{lrccccc} 
\toprule
    & Size & Humanities & Social sciences & STEM & Other & Average       \\ 
\midrule
%GPT3                  & 175B   & 40.8       & 50.4            & 36.7 & 48.8  & 43.9          \\ \midrule
%MTNLG                  & 530B  & 58.3       & 60.0            & 42.8 & 55.4  & 52.2          \\ \midrule
%\multirow{3}{*}{PaLM} & 8B    & 25.6       & 24.1       & 23.8 & 24.1   & 25.3          \\
%& 62B   & 59.5       & 62.7         & 41.9 & 55.8  & 53.7          \\
%& 62B-cont   & -      & -           & - & - & \textbf{61.2}\\
%\midrule
%\multirow{2}{*}{LLaMA-2} & 7B  & 42.9  & 51.2 & 36.4& 52.2  & 45.3 \\
%& 13B  & 52.8  & 62.6 & 44.1 & 66.1 & 54.8  \\
%& 34B & 59.4 & 71.8 & 52.1 & 69.2 & 62.6 \\
%\midrule
%\multirow{2}{*}{Baichuan-2} & 7B  & -      & -           & - & -   & 54.2 \\
%& 13B  & -      & -           & - & -   & 59.2 \\
%\midrule
%QWEN & 14B & 60.9 & 76.2 & 59.4 & 71.8 & \textbf{66.3} \\
%\midrule
%Mistral & 7B & -& -& -& -& 60.1 \\
%\midrule
%Gemma & 7B & -& -& -& -&  64.3 \\
%\midrule
\ours{} & 15B & 69.2 & 74.1 & 53.4 & 67.5 & 64.2 \\ 
\bottomrule
\end{tabular}
\caption{\label{tab:mmlu} Per-category breakdown accuracy for MMLU }
\end{table}

\end{document}